\pdfoutput=1

\documentclass[11pt]{article}

\usepackage[]{acl}

\usepackage{times}
\usepackage{latexsym}

\usepackage[T1]{fontenc}

\usepackage[utf8]{inputenc}

\usepackage{microtype}

\usepackage{inconsolata}

\usepackage{amsthm}
\usepackage{amsmath}
\usepackage{amssymb}
\usepackage{caption}
\usepackage{subcaption}
\usepackage{enumerate}
\usepackage{multirow}
\usepackage{graphicx} 
\usepackage{url}
\usepackage{booktabs}
\usepackage{color}
\usepackage{algorithm}
\usepackage{algorithmic}
\usepackage{makecell}
\usepackage{tcolorbox}
\usepackage{enumitem}

\usepackage{cleveref}
\crefname{section}{§}{§§}
\Crefname{section}{§}{§§}

\title{Can LLMs Reason with Rules? \\Logic Scaffolding for Stress-Testing and Improving LLMs}

\author{Siyuan Wang\textsuperscript{\rm 1}, Zhongyu Wei\textsuperscript{\rm 1}\thanks{~~Corresponding author}, Yejin Choi\textsuperscript{\rm 2,4}, Xiang Ren\textsuperscript{\rm 3,4} \\
\textsuperscript{\rm 1}Fudan University,
\textsuperscript{\rm 2}University of Washington, \\
\textsuperscript{\rm 3}University of Southern California, \textsuperscript{\rm 4}Allen Institute for Artificial Intelligence \\
\texttt{wangsy18@fudan.edu.cn} \\
}

\begin{document}
\maketitle
\begin{abstract}
Large language models (LLMs) have achieved impressive human-like performance across various reasoning tasks. However, their mastery of underlying inferential rules still falls short of human capabilities. To investigate this, we propose a logic scaffolding inferential rule generation framework, to construct an inferential rule base, ULogic, comprising both primitive and compositional rules across five domains. Our analysis of GPT-series models over a rule subset reveals significant gaps in LLMs' logic understanding compared to human performance, especially in compositional and structural complex rules with certain bias patterns. We further distill these rules into a smaller-scale inference engine for flexible rule generation and enhancing downstream reasoning. Through a multi-judger evaluation, our inference engine proves effective in generating accurate, complex and abstract conclusions and premises, and improve various commonsense reasoning tasks. Overall, our work sheds light on LLMs' limitations in grasping inferential rule and suggests ways to enhance their logical reasoning abilities~\footnote{Code and data are available at \url{https://github.com/SiyuanWangw/ULogic}.}.
\end{abstract}

\section{Introduction}
\begin{figure}[!t]
    \centering
    \includegraphics[width=0.99\columnwidth]{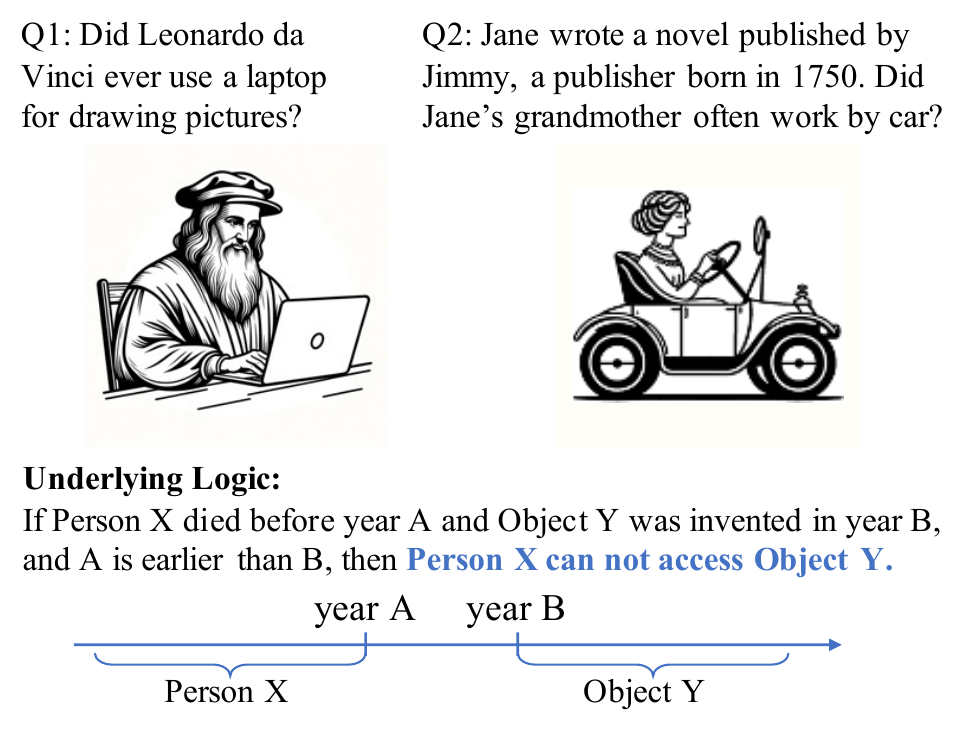}
    \caption{The underlying logic to answer Q1 and Q2.}
    \vspace{-1mm}
    \label{sec1:intro_case}
    \vspace{-3mm}
\end{figure}

``\textit{Did Leonardo da Vinci ever use a laptop for drawing pictures?}'' Large language models can swiftly and confidently respond ``\textit{No}"~\cite{geva2021did,wang2023plan}, demonstrating impressive reasoning ability that rivals human~\cite{openai2023gpt4,ouyang2022training}. However, when posed with more obscure questions, such as question Q2 in Figure~\ref{sec1:intro_case}, LLMs are prone to exhibit uncertainty and errors. 
This inconsistency raises concerns about whether LLMs grasp the underlying logic of matters as proficiently as humans~\cite{wason1968reasoning} (see ``\textit{underlying logic}" in Figure~\ref{sec1:intro_case}) and highlights challenging reasoning situations (like Q2) where current LLMs might struggle.

\begin{figure*}[th!]
    \centering
    \includegraphics[width=0.98\textwidth]{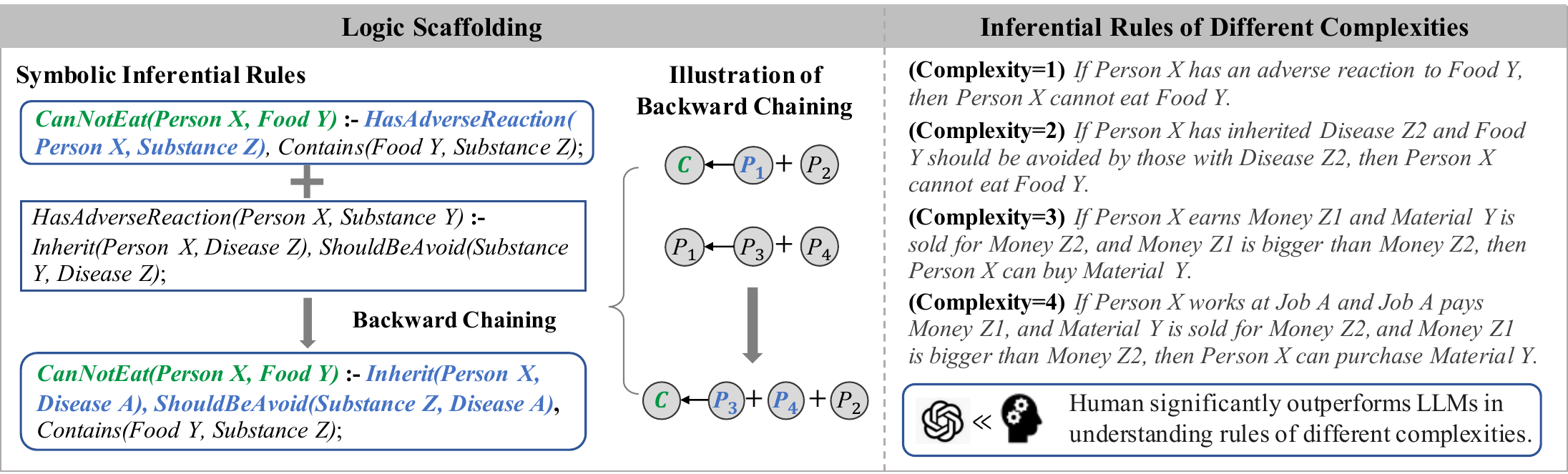}
    \caption{Logic scaffolding uncovers challenging reasoning space for LLMs. (Complexity refers to the rule length.)}
    \label{sec1:intro_performance}
\end{figure*}

Humans naturally abstract underlying logic (\textit{e.g.}, inferential rules) from extensive real-world observations~\cite{barwise1993everyday}, which is beneficial for addressing diverse reasoning situations. An inferential rule is typically defined as a premise with a set of facts (e.g., ``\textit{Person X died before ... earlier than B}'') leading to a conclusion (e.g., `\textit{`Person X cannot access Object Y}'')~\cite{boghossian2014inference}. Grasping this rule enables the deduction that a person cannot access an object invented posthumously.
This work utilizes symbolic logic as a \textit{scaffold} to generate challenging reasoning situations for GPT-series LLMs, as shown in Figure~\ref{sec1:intro_performance}. We observe a discernible gap between LLMs and humans in understanding inferential rules, especially rules with complex premises.

However, collecting such inferential rules at scale presents a major challenge. Previous work mainly relies on manual curation~\cite{sap2019atomic,sinha2019clutrr} or inductive logic programming~\cite{qu2020rnnlogic}, which are either labor-intensive or limited in diversity. Besides, manually crafted rules often appear simple and overly specified, struggling to move beyond basic intuition or generalize across diverse situations. For example, the rule \emph{If X runs out of steam, then X becomes tired} from~\citet{sap2019atomic} has only one premise fact and narrowly specifies exhaustion. 

To this end, we introduce \textbf{L}ogic scaff\textbf{O}lding \textbf{I}nferential \textbf{R}ule g\textbf{E}neration (\textbf{LOIRE}), a framework to generate inferential rules of different complexities.
LOIRE operates in two stages: primitive rule generation and rule composition.
Initially, we define ``primitive rules'' to describe abstract objects like \texttt{Person} and \texttt{Food}, 
and ensure they cannot be decomposed into simpler rules,
facilitating broad generalization and easy generation. We then incorporate GPT-4's generative capability and human expertise to generate primitive rules with high confidence. 
This process, consistently guided by symbolic logic, involves GPT-4 drafting potential conclusions in various domains, and forming premises with one or more facts. We ensure rules' logical soundness through the model's self-critique and human manual verification. 
In the second stage, we apply backward chaining~\cite{gallaire2012logic, al2015comparison} upon these primitive logical rules to automatically construct compositional rules of varied lengths and structures at scale. 

Using this framework, we construct \textbf{ULogic}, an inferential rule base with around $8,000$ primitive rules and over $6,000$ compositional rules. 
These rules span five key domains: object affordance, accessibility, interaction, location, and person's need. We hope ULogic will serve as a valuable resource, facilitating the assessment of LLMs' proficiency in underlying logic and enhancing flexible rule generation and downstream reasoning.
We utilize ULogic to create an entailment probing task with a comprehensive and robust evaluation strategy, to assess LLMs' grasp of inferential rules against human performance. Our analysis of GPT-series LLMs, including GPT-4, GPT-3.5-Turbo and GPT-3.5-Turbo-Instruct,
indicates they have a basic understanding of inferential rules but fall short of human proficiency, especially in rules with complex premises. 
Specifically, all models struggle more as the compositional complexity increases. While GPT-4 performs consistently on verbalized and symbolic rules, the other models sharply degrade on symbolic rules. Additionally, all models exhibit disparities on various rule structures with Disjunctive-Transitive rules posing the greatest challenges. Moreover, these LLMs display notable polarity biases with GPT-4 showing a necessary bias, underscoring areas for improvement.

We further distill crafted inferential rules into a smaller-scale inference engine for flexible rule generation and downstream reasoning. We design three tasks: conclusion generation, premise completion and premise generation, to construct an instruction-tuning dataset for inferential rule distillation. Experimental results through a multi-judger evaluation mechanism incorporating automatic metrics, LLM evaluators and human preferences show that our inference engine possesses the ability for these three tasks. 
It outperforms GPT-3.5-Turbo across all dimensions of three tasks and even surpasses GPT-4 in generating more complex and abstract rules. Moreover, it can generate logical rules that enhance downstream commonsense reasoning.

\section{Logic Scaffolding for Inferential Rule Generation}
\begin{figure*}[t!]
    \centering
    \includegraphics[width=0.98\textwidth]{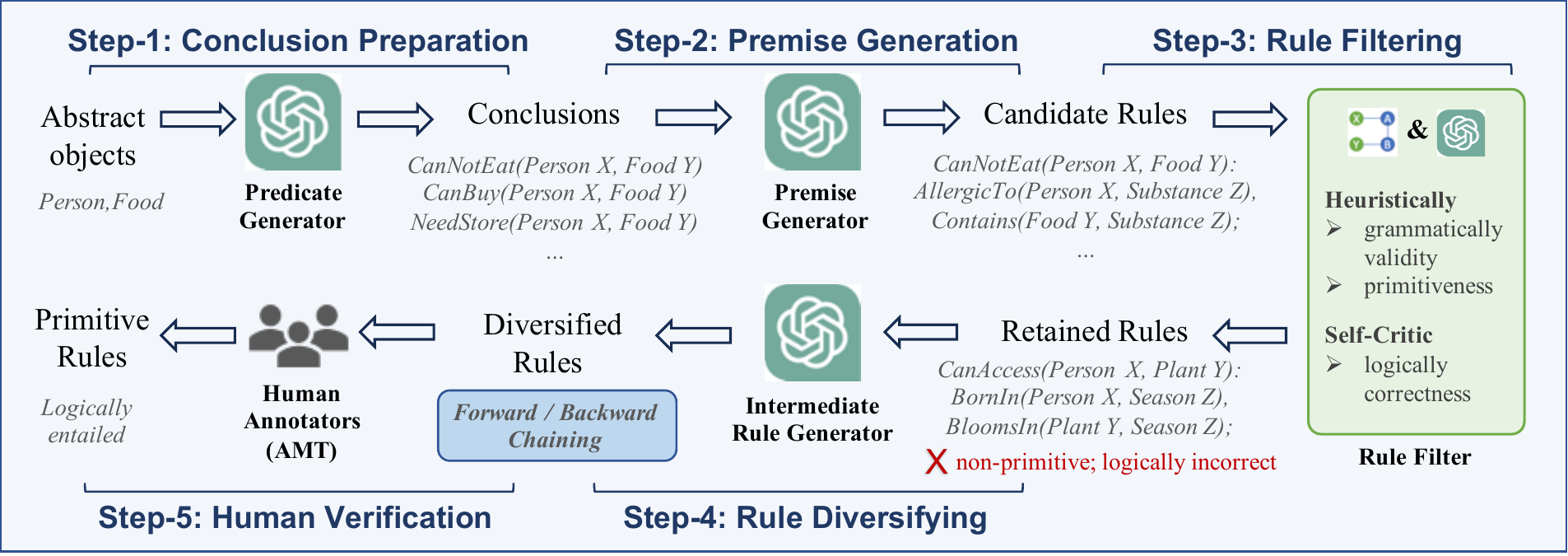}
    \caption{The pipeline of primitive rule generation.}
    \label{sec2:generattion_pipeline}
\end{figure*}
\subsection{Preliminary of Inferential Rules} 
To better control the generative capability of LLMs for rule generation, we focus on \emph{if-then} inferential rules with variables, that can be easily expressed as symbolic logic~\cite{novak2006logical}. 
An inferential rule describes a logical implication from a premise (a set of facts) to a conclusion (a specific fact), where each fact is a predicate expression with two variables, and each variable has a designated variable type. For each rule, we employ logic scaffolding which first generates its symbolic expression to consistently guide its verbalized form.

We utilize Prolog~\cite{apt1997logic} to formulate symbolic rules as \texttt{Conclusion:-Premise}, where \texttt{:-} indicates the logical implication. 
For example, 
\begin{align}
    \emph{\text{CanNot}}&\emph{\text{Eat}}(\emph{\text{Person}}\ X, \emph{\text{Food}}\ Y) \texttt{:-} \nonumber \\ 
    &\emph{\text{AllergicTo}}(\emph{\text{Person}}\ X, \emph{\text{Substance}}\ Z), \nonumber \\  
    &\emph{\text{Contains}}(\emph{\text{Food}}\ Y, \emph{\text{Substance}}\ Z). 
\end{align} 
The left-hand side is the conclusion and the right hand lists premise facts connected by commas. ``CanNotEat'', ``AllergicTo'' and ``Contains'' are predicate verbs while \texttt{Person, Food, Substance} are variable types of variables (X, Y, Z).
This symbolic rule can be verbalized as:
\emph{If Person X is allergic to Substance Z and Food Y contains Substance Z, then Person X cannot eat Food Y}.

\vspace{2mm}
\noindent\textbf{Primitive Rule} We aim to generate primitive rules for further compositions and potential generalization. We formally define primitive rules as follows: (1) they concern abstract objects, like \texttt{Person} and \texttt{Food}, rather than specific instances, and their common properties; (2) they cannot be decomposed into simpler rules.
Inspired by superordinate objects such as  \texttt{instrument, fruit, tool} from \citet{rosch1975family}, we assemble a collection of abstract objects. We first identify the most common tail nodes of ``IsA'' relations from ConceptNet~\cite{speer2017conceptnet}. For those nodes that are still fine-grained, we further seek their general hypernyms by searching ConceptNet and WordNet~\cite{miller1995wordnet}. We totally gather a list of 32 most common abstract objects for primitive rule generation, with 18 common properties generated by prompting GPT-4, as detailed in Appendix~\ref{appen:abstract_objects}.

\subsection{Primitive Rule Generation Pipeline}
\label{chap_pipeline}
The pipeline of primitive rule generation is illustrated in Figure~\ref{sec2:generattion_pipeline}, consisting of five steps. First, we randomly select two abstract objects, and generate potential predicates between them to form conclusions. GPT-4 is prompted to generate corresponding feasible premises with both single and multiple facts, thereby constructing candidate primitive rules. We then apply heuristic methods to filter invalid and non-primitive rules, and utilize GPT-4 to select the rules it deems logically correct. We further diversify rule predicates via backward/forward chaining~\cite{urbani2011querypie, shindo2021neuro} with generated single-fact rules, and filter excessively repetitive rules. 
Finally, the diversified rules undergo manual verification to ensure the final set of high-confidence primitive rules.

\vspace{1.5mm}
\noindent\textbf{Step-1: Conclusion Preparation}
From the set of abstract objects, we select any two,
e.g., \texttt{Person} and \texttt{Food}, and prompt GPT-4 to generate potential predicates connecting them as conclusions, e.g., \emph{CanEat(Person X, Food Y)}. We attempt every possible pairing of two, where the selected objects can be identical.
For each pair of objects, \{object$_1$\} and \{object$_2$\}, we aim to generate conclusions across five domains: \{object affordance, accessibility, interaction, location and person's need\}, thereby covering diverse scenarios. Explanations and example rules of these domains are listed in Appendix~\ref{rule_domain}. 
The prompt for conclusion preparation about affordance is below. Besides, we negate the generated predicates to yield both positive and negative conclusions, e.g., \emph{CanNotEat(Person X, Food Y)}, across object affordance, accessibility, and interaction domains, building a complete rule set.
\begin{center}
\begin{tcolorbox}[colback=blue!5!white,colframe=black!55!black,width=0.49\textwidth,title={Prompt for Conclusion Preparation}]
\small
According to commonsense knowledge in reality, please list 5 predicates between the given two objects to describe the \{\underline{object affordance}\}. \\
Examples: \\
Object: Show, Artwork  \\
Predicate: CanBeAdaptedFrom(Show X, Artwork Y) \vspace{2mm}\\
\textbf{Object}: \{\underline{object$_1$}\}, \{\underline{object$_2$}\} \\
\textbf{Predicate}:
\end{tcolorbox}
\end{center}

\vspace{1.5mm}
\noindent\textbf{Step-2: Premise Generation}
\label{chap_rule_generation}
Guided by a symbolic conclusion, we prompt GPT-4 to generate its premises in both symbolic and verbalized forms for better controllability. This process involves the logit bias setting, motivating premises to describe relationships between abstract objects and their properties.
Specifically, premises are generated under the constraint of logit bias, increasing the likelihood of these objects and properties appearing in the output. For each conclusion, we create both single-fact and multi-fact premises to yield candidate rules of varying lengths. We tailor instructions and demonstrations for each domain to prompt GPT-4 for premise generation exploring different possibilities, as detailed in Appendix~\ref{rule_generation_prompts}.

\vspace{1.5mm}
\noindent\textbf{Step-3: Rule Filtering}
\label{chap_rule_filtering}
After over-generating candidate primitive rules, we first design several heuristic methods to filter grammatically invalid or non-primitive rules, based on their symbolic forms. For grammatically validity, we check whether the variables in premises form a connected graph from node ``X'' to node ``Y'', as in Appendix~\ref{appen:grammar_validity}.
Regarding primitiveness, we exclude rules with non-primitive variable types or those comprising more than 3 premise facts.
Besides, we eliminate trivial rules that both contain negative words in their premise and conclusion, e.g., \emph{CanNotEat(Person X, Food Y):- CanNotAccess(Person X, Food Z)}.

Directly generating logically correct rules is  challenge. Thus we further adopt a self-critic strategy~\cite{gou2023critic} where GPT-4 critiques the accuracy of its self-generated rules in a verbalized format, and provides explanations of its judgments. When prompting GPT-4, we include two demonstrations featuring both correct and incorrect rules to mitigate label bias. These demonstrations vary across different domains. An example prompt for object affordance is in Appendix~\ref{rule_filtering_prompts}.

\begin{figure}[h!]
    \centering
    \includegraphics[width=0.98\columnwidth]{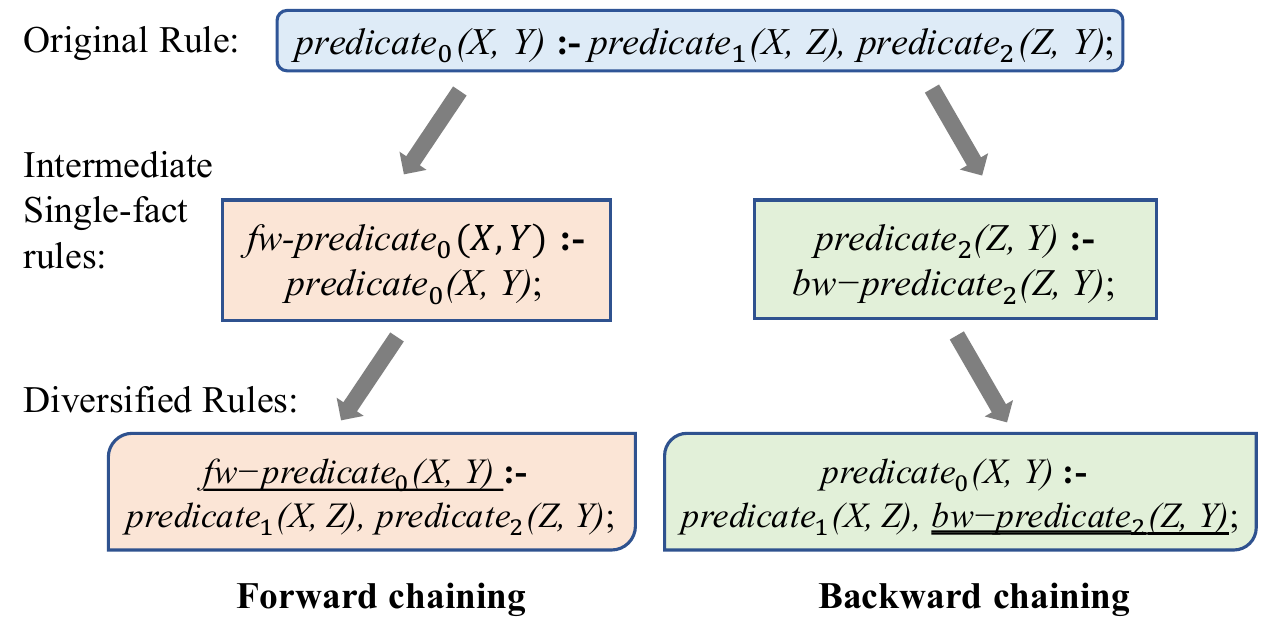}
    \caption{The forward and backward chaining process for diversifying rules.}
    \label{sec2-2:rule_diverse}
\end{figure}

\vspace{1.5mm}
\noindent\textbf{Step-4: Rule Diversifying}
\label{chap_rule_diverse}
To increase the variety of rule expressions, we diversify  predicates while maintaining its logical accuracy. 
Based on symbolic rules, we respectively apply forward and backward chaining algorithms to their conclusion and premise with generated single-fact rules, as shown in Figure~\ref{sec2-2:rule_diverse}.
In forward chaining, we take the conclusion as a new premise to generate an intermediate single-fact rule, subsequently substituting the original conclusion with this newly derived conclusion.
In backward chaining, a premise is taken as a conclusion to create an intermediate single-fact rule, and replace the original premise with the new-generated one. Intermediate single-fact rules are also generated through Step-2 and 3. Each original rule undergoes one forward and one backward chaining  to derive two diversified rules.

\noindent\textbf{Step-5: Human Verification}
To obtain more reliable inferential rules, we utilize Amazon Mechanical Turk (AMT) to recruit human annotators for manual verification. For each rule, three annotators are asked to assess 
the clarity and comprehensibility of its premise and conclusion,
as well as the logical entailment from the premise to the conclusion. Only the rules unanimously validated by all three annotators are preserved. The AMT template for human verification and the overall rates of rule acceptance are listed in Appendix~\ref{human_verify_template}.

\subsection{Rule Composition}
\begin{figure}[!t]
    \centering
    \includegraphics[width=0.96\columnwidth]{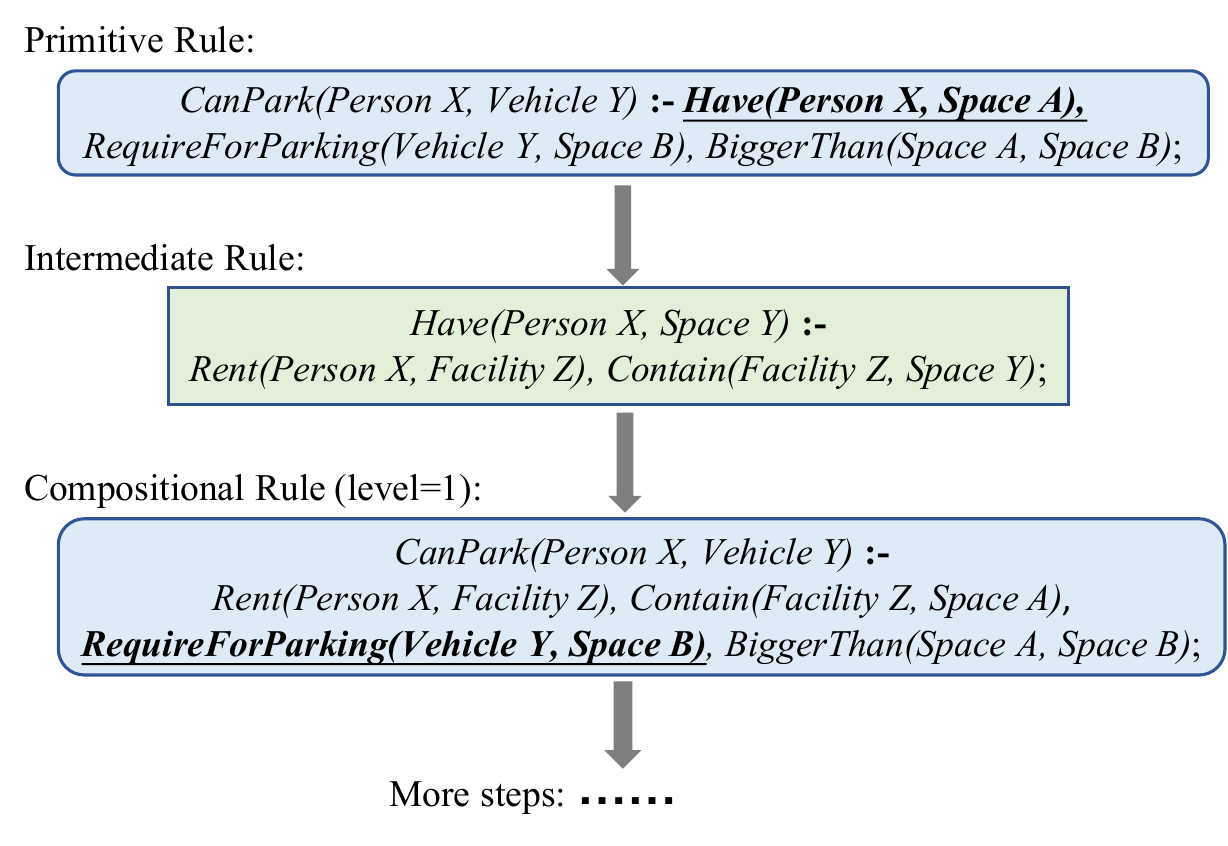}
    \caption{Illustration of one backward chaining step.}
    \label{sec2-2:rule_composition}
    \vspace{-1mm}
\end{figure}
We create more compositional rules by applying backward chaining upon primitive rules with different chaining steps. In each step, we select a premise fact from the current rule as a conclusion, deriving a new primitive rule that describes its multi-fact premise.
This selected fact is then replaced with the newly generate premise.
This process is iteratively conducted 1 to 3 times, creating rules with varying compositional levels (1 to 3). An example of one backward chaining step is shown in Figure~\ref{sec2-2:rule_composition}.
The intermediate primitive rules used in backward chaining are generated via the pipeline described in Sec.~\ref{chap_pipeline}, thus also contributing to our primitive rule set. 
As the composition of logically correct sub-rules is also logically correct, there is no need to verify these compositional rules separately.


\subsection{Rule Statistics}
\renewcommand{\thesubfigure}{\alph{subfigure}}
\begin{figure*}[ht!]
    \centering
    \begin{minipage}[b]{.65\linewidth}
    \begin{subfigure}{0.5\textwidth}
        \centering
        \includegraphics[width=\textwidth]{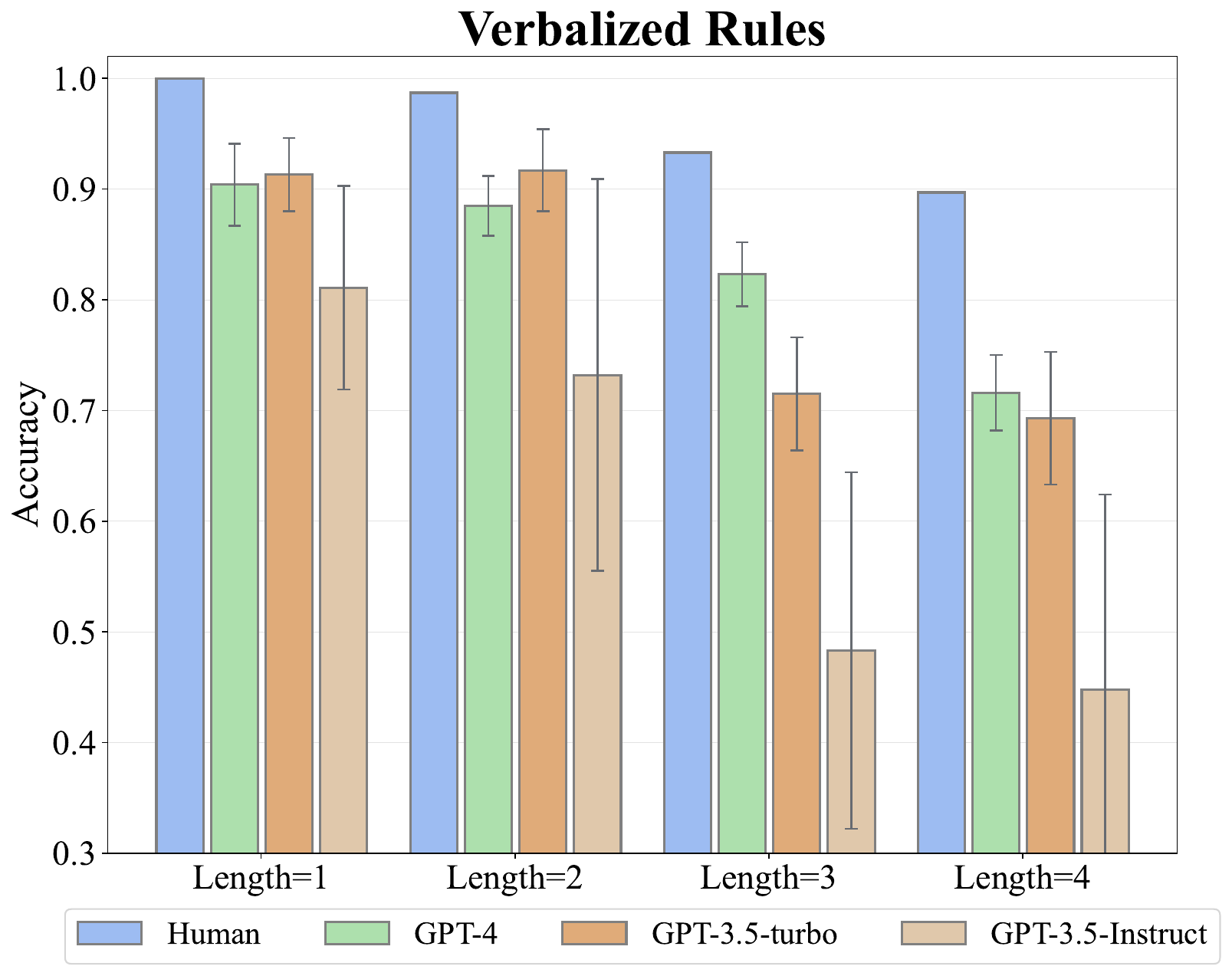}
        \caption{}
        \label{length_analysis}
        \vspace{-1mm}
    \end{subfigure}
    \hfill
    \begin{subfigure}{0.448\textwidth}
        \centering
        \includegraphics[width=\textwidth]{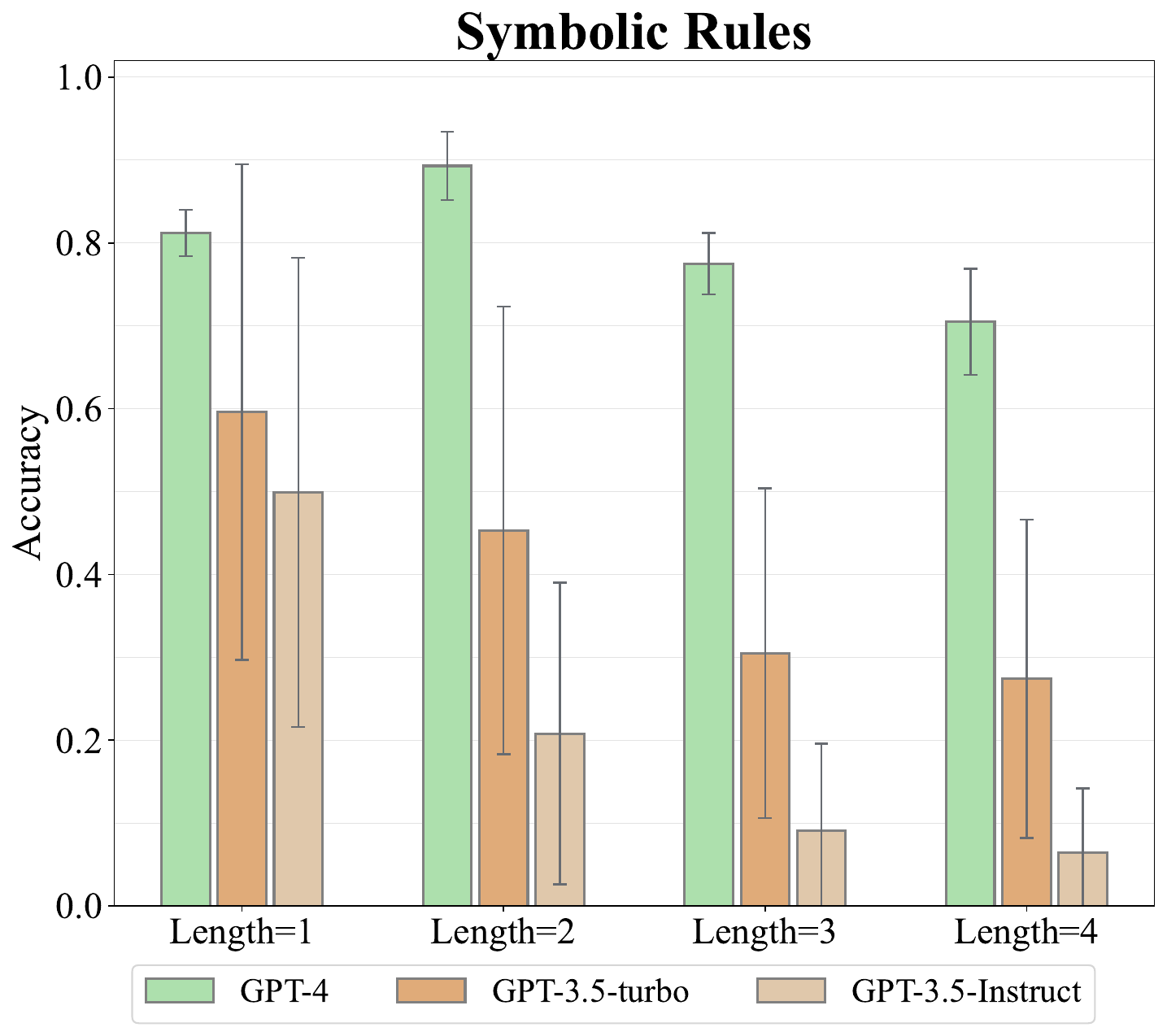}
        \caption{}
        \label{symbolic_analysis}
         \vspace{-1mm}
    \end{subfigure}
    \vspace{-2mm}
    \caption{Probing results across varied lengths.}
    \label{overall_analysis}
    \end{minipage}
    \hfill
    \begin{minipage}[b]{.33\linewidth}
        \begin{subfigure}{\textwidth}
        \includegraphics[width=\textwidth]{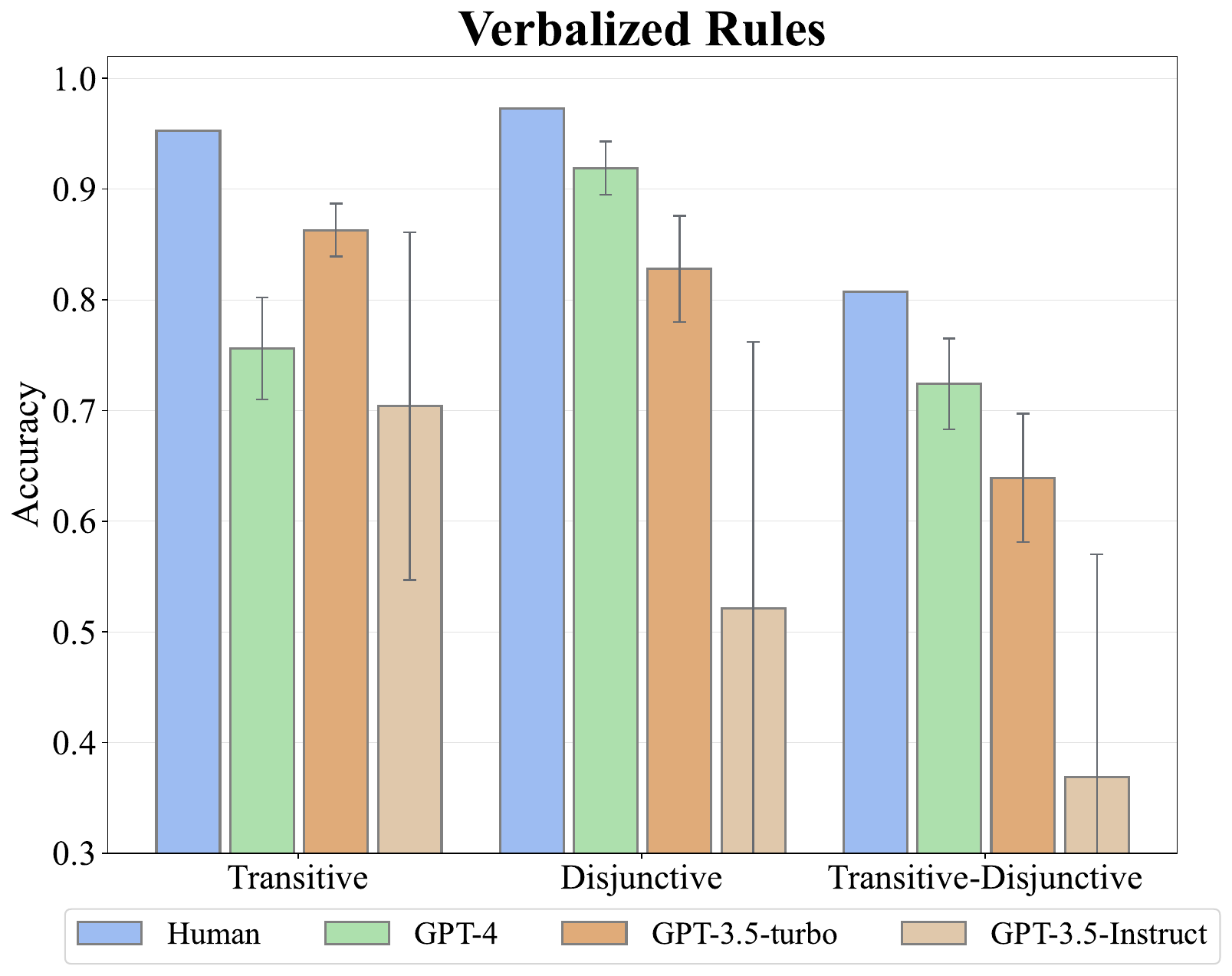}
        \vspace{1mm}
        \end{subfigure}
        \vspace{-4mm}
        \caption{Result of varied structures.}
        \label{structure_analysis}
    \end{minipage}
    \vspace{-1mm}
\end{figure*}
Using Logic-RGC framework, we construct an inferential rule base ULogic comprising $14,647$ rules, with $7,967$ primitive and $6,680$ compositional ones. These rules span five key domains: object affordance, accessibility, interaction, location and person's need.
They vary in compositional depth from 0 to 3, with rule lengths ranging from 1 to 6. Detailed statistics are in Table~\ref{sec2.4_statistics}.
\begin{table}[th!]
    \centering
    \resizebox{0.48\textwidth}{!}{
    \begin{tabular}{c|ccccc|c}
        \toprule
        Domain & Aff. & Acc. & Int. & Loc. & Need & \ \ \ Total \ \ \ \\
        \midrule
        \multicolumn{6}{c}{ Primitive rules } & 7,967 \\
        \midrule
        Single-fact & 328 & 513 & 440 & 194 & 87 & 1,562 \\
        Multi-fact & 387 & 638 & 2,527 & 166 & 128 & 3,846 \\
        Intermediate & 417 & 590 & 1,286 & 165 & 101 & 2,559 \\
        \midrule
        \multicolumn{6}{c}{ Compositional rules } & 6,680\\
        \midrule
        Comp.=1 & 322 & 675 & 936 & 111 & 91 & 2,135 \\
        Comp.=2 & 199 & 773 & 744 & 100 & 136 & 1,952 \\
        Comp.=3 & 229 & 1052 & 896 & 217 & 199 & 2,593 \\
        \bottomrule
    \end{tabular}
    }
    \caption{Statistics of constructed rule base. Aff., Acc.,  Int., Loc., Comp. are abbreviations of Affordance, Accessibility, Interaction, Location and Compositionality.}
    \label{sec2.4_statistics}
\end{table}

\section{Assessing LLMs' Proficiency in Capturing Inferential Rules}
We utilize ULogic for a systematic evaluation of LLMs' proficiency in underlying logic compared to human competence. Specifically, we select a high-quality probing subset with 1,104 diverse rules from our rule base~\footnote{These high-quality probing rules are verified by authors, covering various lengths, polarities and structures.}, and create a binary entailment classification task for assessing whether LLMs capture the entailment within inferential rules. 

\subsection{Analysis Setup}
Considering LLMs' sensitivity to various input formulations and shortcut biases, we design a comprehensive and robust assessment mechanism to ensure reliable analysis. For each inferential rule, we convert it into five distinct probing questions to mitigate template bias, as summarized in Appendix~\ref{appen:probing_templates}. We report the average accuracy and variance (the error line of each bar) across five templates. Besides, we adopt a two-shot chain of thought (CoT) prompting strategy~\cite{wei2022chain} requiring the model to generate a rationale after presenting its answer, using ''and also explain why.'' We include one correct rule and one incorrect rule in the two demonstrations to minimize label bias. 

Following the Law of Non-Contradiction~\cite{priest2006law}, the propositions "If X, then Y" and "If X, then not Y" are mutually exclusive that cannot both be true at the same time. To enhance the reliability of our probing, we flip each rule by negating its conclusion, and simultaneously probe both the original rule and its flipped version. 
A rule is accurately classified only if the original rule is affirmed (True/Right/Yes) and its flipped counterpart is negated (False/Wrong/No), as shown below. A specific example is in Appendix~\ref{appen:both_side_probing}. This dual-sided probing is applied to both human and LLMs.
\begin{table}[!h]
    \vspace{-4mm}
    \resizebox{0.49\textwidth}{!}{
    \begin{tabular}{m{5.7cm}cp{1cm}}
    \toprule 
    \emph{If Premise, then Conclusion$\_$original.} & True/Right/Yes \\
    \emph{If Premise, then Conclusion$\_$flipped.} & False/Wrong/No \\
    \bottomrule
    \end{tabular}
    }
    \vspace{-3mm}
\end{table}

\subsection{Empirical Analysis}
\label{sec3_2_analysis}
We conduct analysis on GPT-series LLMs, including GPT-4, GPT-3.5-Turbo and GPT-3.5-Turbo-Instruct\footnote{Substituting the now-deprecated Text-davinci-003.}, aiming to investigate LLMs' proficiency of inferential rules against human performance by exploring the following questions. The human performance is obtained by asking AMT annotators whether the input rule is logical correct with high probability. Each performance presented in following bar charts is calculated based on 150 instances randomly sampled from our probing subset.

\vspace{1mm}
\noindent\textbf{(1) How does model performance vary with increasing compositional complexity?} We conduct rule probing in terms of different compostional lengths, as illustrated in Figure~\ref{length_analysis}. ``Length=1,2,3,4'' respectively denote rules with 1$\sim$4 facts in their premises.
The analysis of different compostional depths is also provided in Appendix~\ref{appen:rule_depth_probing}. 
They both reveal that as compositional complexity increases, the performance of both human and all models drops. The primary reason is that compositional complex rules typically necessitate the aggregation of multi-step reasoning, which escalates higher-order relationships understanding and exponential error accumulation with each additional step~\cite{dziri2023faith}.
Besides, there is a persistent performance gap between all models and human, particularly pronounced with compositional complex rules, suggesting significant potential for enhancement in this area.

\vspace{1mm}
\noindent\textbf{(2) Are LLMs proficient in capturing both symbolic and verbalized rules?}
We further analyze the performance of LLMs on symbolic rules (see Figure~\ref{symbolic_analysis}), and compared it to the verbalized result in Figure~\ref{depth_analysis}.
We observe that GPT-4 achieves consistent performance on verbalized and symbolic rules, whereas GPT-3.5-Turbo and GPT-3.5-Instruct sharply degrade on symbolic rules. This suggests that 
the GPT-3.5 series may have limitations in generalizing across varied types of linguistic structures beyond natural language, whereas GPT-4 likely have undergone specific optimizations for symbolic interpretations.

\vspace{1mm}
\noindent\textbf{(3) Are there performance disparities among models concerning different rule structures?} Our generated multi-fact rules (Length $> 1$) have three intrinsic structures: Transitive, Disjunctive and Disjunctive-Transitive.
Specific illustrations and examples of each structure are detailed in Appendix~\ref{appen:rule_structure}.
Figure~\ref{structure_analysis} shows that Disjunctive-Transitive rules pose greater challenges compared to Transitive and Disjunctive ones, especially for GPT-3.5-Turbo and GPT-3.5-Instruct. We hypothesize that this discrepancy stems from increased compositional complexity and LLMs' insufficient learning of logical structures in natural language.

\begin{figure}[th!]
    \centering
    \begin{subfigure}{0.235\textwidth}
        \centering
        \includegraphics[width=\textwidth]{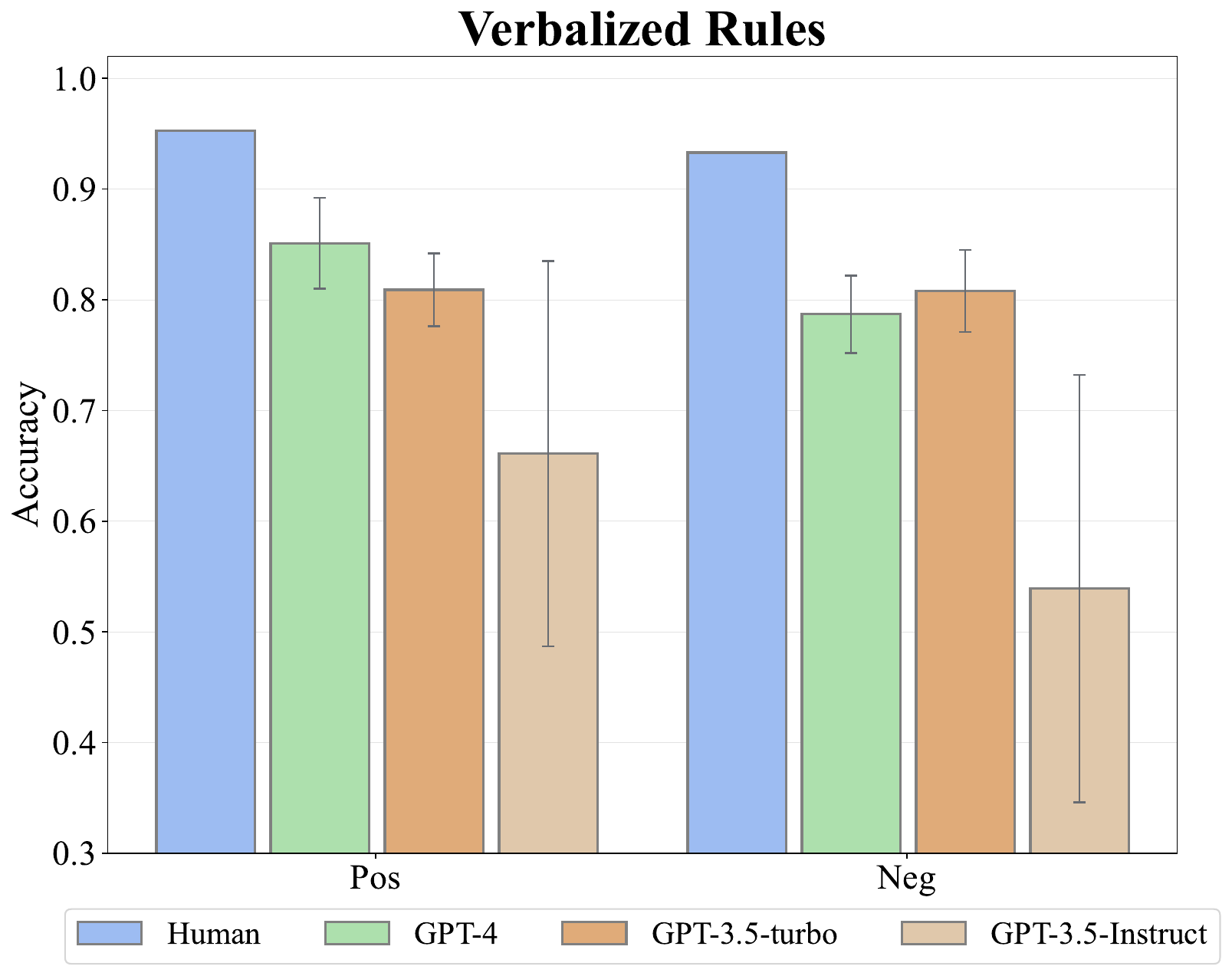}
        \subcaption{Answer-Explain strategy.}
        \label{polarity_analysis}
        \vspace{-1mm}
    \end{subfigure}
    \begin{subfigure}{0.21\textwidth}
        \centering
        \includegraphics[width=\textwidth]{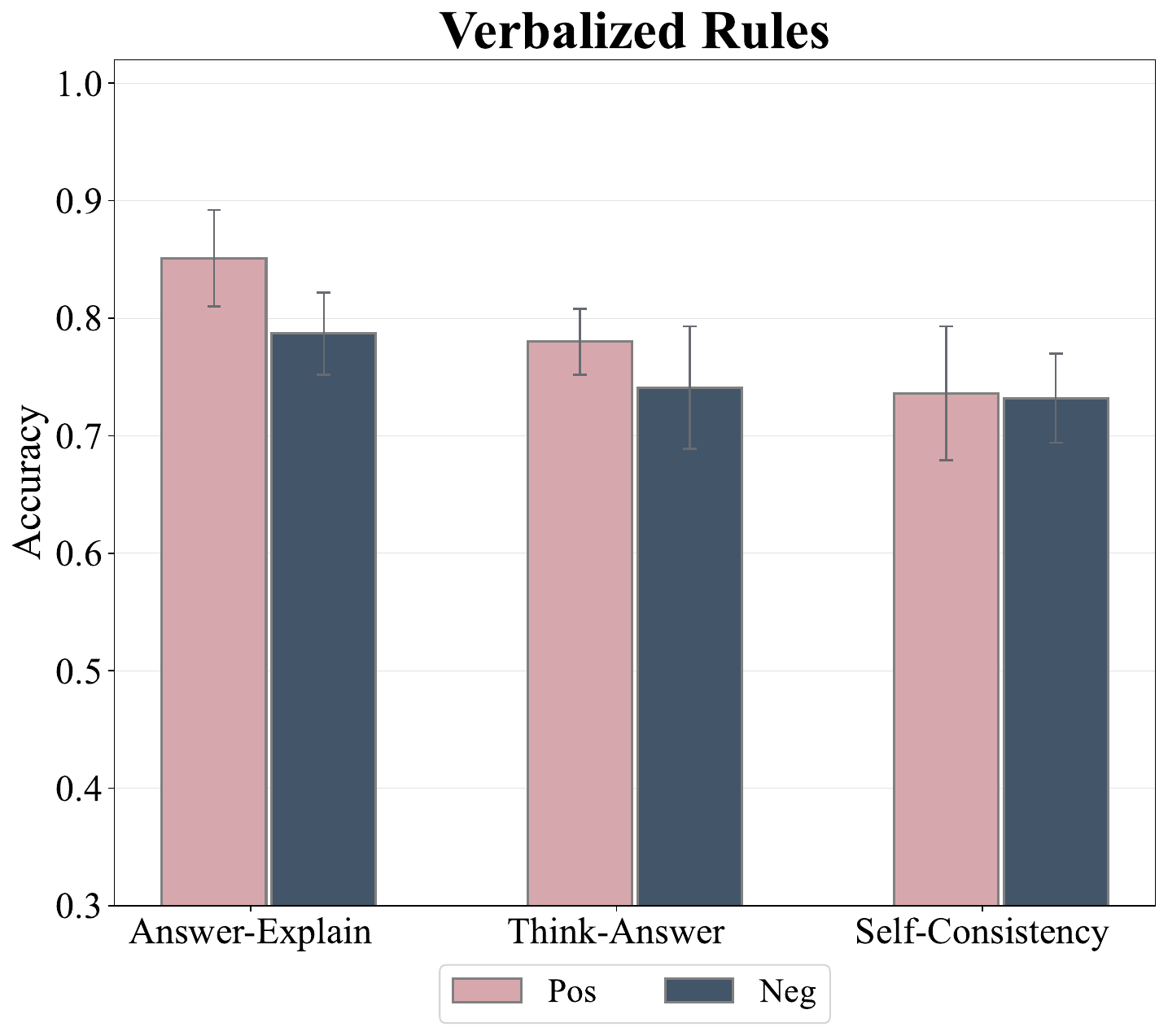}
        \subcaption{Various CoT strategies.}
        \label{cot_comparison}
        \vspace{-1mm}
    \end{subfigure}
    \caption{Rule Polarity Comparison.}
    \label{overall_polarity_comparison}
\end{figure}

\vspace{1mm}
\noindent\textbf{(4) Do LLMs exhibit a polarity bias over inferential rules?}
Our inferential rules contain both positive and negative conclusions. We conduct a comparative analysis of polarity discrepancy, as shown in Figure~\ref{polarity_analysis}.
GPT-4 and GPT-3.5-Instruct exhibit a pronounced positive bias, performing better on rules with positive conclusions. This bias may originate from the imbalanced distribution of LLMs' training data~\cite{garg2022identifying}, with a higher proportion of positive statements.
We further explore different CoT strategies with GPT-4: (1) first answer then explain (\emph{Answer-Explain}), (2) first think then answer (\emph{Think-Answer}), (3) self-consistently think then answer (\emph{Self-Consistency})~\cite{wang2022self}. Various CoT prompts are listed in Appendix~\ref{appen_cot_prompts}.
Figure~\ref{cot_comparison} shows that although advanced CoT strategies can mitigate the positive bias, they adversely impact the performance on rules with both positive and negative conclusions. 
\renewcommand{\thesubfigure}{\alph{subfigure}}

\noindent\textbf{(5) Why does GPT-4 significantly underperform GPT-3.5-Turbo on transitive rules?}
Previous analysis shows that GPT-4 generally outperforms or equals the other models, but this superiority disappears on transitive rules, as evidenced in Figure~\ref{structure_analysis}. 
We investigate this question in Appendix~\ref{necessary_bias}, which reveals that GPT-4 exhibits a ``necessary bias'' that tend to consider all necessary conditions reaching a conclusion, avoiding definite judgement. This conservative style may come from LLMs' preference alignment during Reinforcement Learning with Human Feedback~\cite{ouyang2022training}.

Overall, GPT-4 performs best in grasping inferential rules. But compared to human performance, there still remains substantial room for improvement across all models, especially in highly compositional, symbolic and structural complex rules. Besides, all models tend to exhibit a polarity bias towards rules with positive conclusions with GPT-4 also showing a necessary bias. These findings suggest potential areas for future enhancements.


\section{Rule Distillation as Inference Engine}
\subsection{Instruction Dataset \& Model Tuning}
For flexible rule generation and benefiting downstream reasoning, we distill our crafted rules into a smaller-scale inference engine as illustrated in Appendix~\ref{appen:instruction_tuning_pipeline}. We tailor three tasks: conclusion generation, premise completion and premise generation, to construct an instruction-tuning dataset for inferential rule distillation. The detailed definitions of these tasks are also described in Appendix~\ref{appen:instruction_tuning_pipeline}.

We gather all primitive rules and partial compositional rules to formulate the instruction-tuning dataset, as compositional rules are constructed from primitive ones. We take 10,703 rules for training and 943 for testing. Altogether, we create 39,887 instances for instruction tuning, including 10,703, 18,500 and 10,684 for conclusion generation, premise completion and premise generation. We have 3,500 testing instances, divided as 943, 1,614 and 943 for these three tasks.
We use Mistral-7b~\cite{jiang2023mistral} as the backbone model and fine-tune it with our constructed instruction dataset as our inference engine. The training details and demo page can be found in Appendix~\ref{appen:implementation_details}.


\subsection{Rule Generation Evaluation}
We compare our inference engine against GPT-4 and GPT-3.5-Turbo across three tasks to assess rule generation. 
For a fair comparison, we prompt GPT-4 and GPT-3.5-Turbo to simultaneously generate symbolic and verbalized responses, using similar prompts as in Step-2 of Sec.~\ref{chap_rule_generation}. Detailed prompts are in Appendix~\ref{three_task_prompts}.
We introduce a multi-judger \newline evaluation mechanism, incorporating automatic metrics, LLM evaluator and human preference. We evaluate the logical accuracy for conclusion generation and premise completion. 
For premise generation task with a specified number of facts, we generate three potential premises for each conclusion, and compare these premises in accuracy, diversity, complexity and abstractness. Detailed definition of these metrics are described in Appendix~\ref{appen:evaluation_metrics}.

\vspace{1mm}
\noindent\textbf{Automatic Evaluation} 
For automatic accuracy evaluation of three tasks, we calculate BLEU score~\cite{papineni2002bleu} against reference responses. For complexity of premise generation, we assess the average fact number of three generated premises. For diversity, we compute average Self-BLEU~\cite{shu2019generating, tevet2020evaluating} between three generated premises. Specifically, Self-BLEU measures the BLEU score of a generated premise against another, and a high average Self-BLEU indicates low diversity. Abstractness is not easy to evaluate automatically, so we leave it to LLM evaluation. The results are shown in Table~\ref{automatic_result}.
\begin{table}[h]
    \begin{center}
    \resizebox{0.49\textwidth}{!}{
    \begin{tabular}{c|c|c|ccc}
    \toprule
     Task  & Conc Gen & Prem Comp & \multicolumn{3}{c}{Prem Gen} \\
     \midrule
    Metrics & BLEU & BLEU & BLEU & Self-BLEU & Fact Num. \\
    \midrule
    Engine & \bf 0.739 & \bf 0.527 & \bf 0.411 & 0.687 & \bf 3.42 \\
    GPT-4 & 0.414 & 0.179 & 0.149 & \bf 0.805 & 2.58 \\
    GPT-3.5 & 0.338 & 0.248 & 0.084 & 0.739 & 1.72\\
    \bottomrule
    \end{tabular}
    }
    \end{center}
    \caption{Automatic evaluation results. ``Conc Gen'', ``Prem Comp'' and ``Prem Gen'' are abbreviations of conclusion generation, premise completion and generation.}
    \label{automatic_result}
\end{table}

\noindent\textbf{LLM Evaluation}
We adopt GPT-4 as an evaluator to rate the generated responses on a scale from 1 to 3. The criteria of each rating along with examples are provided to the evaluator. Please see Appendix~\ref{llm_evaluation_appen} for detailed prompts. For each task, we select 100 instances for LLM evaluation, ensuring a balance across all domains and all types (including single-fact, multi-fact, intermediate, composition=1$\sim$3 rules) as detailed in Table~\ref{sec2.4_statistics}. The rating results are presented in Table~\ref{llm_result}.
\begin{table}[h!]
    \begin{center}
    \resizebox{0.47\textwidth}{!}{
    \begin{tabular}{c|c|c|cccc}
    \toprule
     Task  & Conc Gen & Prem Comp & \multicolumn{4}{c}{Prem Gen} \\
    \midrule
    Metrics & Acc & Acc & Acc & Div. & Cpx. & Abs. \\
    \midrule
    Engine & 2.44 & \bf 2.78 & 2.34 & 1.89 & \bf 1.62 & \bf 2.43\\
    GPT-4 & \bf 2.53 & 2.72 & \bf 2.77 & \bf 2.64 & 1.40 & 2.32 \\
    GPT-3.5 & 2.38 & 1.57 & 1.91 & 1.72 & 1.06 & 2.30 \\
    \bottomrule
    \end{tabular}
    }
    \end{center}
    \caption{LLM evaluation results. ``Acc'', ``Div.'', ``Cpx.'' and ``Abs.'' are abbreviations of accuracy, diversity, complexity and abstractness.}
    \label{llm_result}
\end{table}

\begin{figure}[h!]
    \centering
    \includegraphics[width=0.48\textwidth]{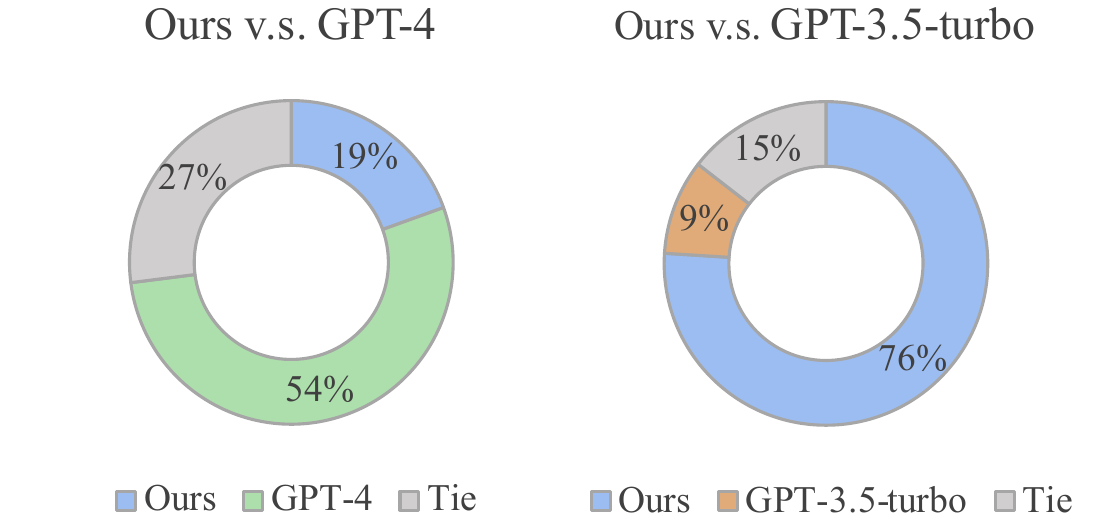}
    \caption{Human comparison results.}
    \label{human_eval}
\end{figure}

\noindent\textbf{Human Evaluation}
To better assess premise generation in line with human value, we further recruit two annotators for each instance to compare their accuracy. We implement a pairwise comparison setting, asking annotators to determine which group of generated premise is more accurate in terms of logical consistency with the given conclusion, commonsense alignment and correctness of fact numbers. The results are shown in Fiure~\ref{human_eval}.

From all evaluation, we can see that our inference engine enables the smaller-scale LLM with the capability for conclusion generation, premise completion and premise generation. 
It performs better than GPT-3.5-Turbo across all metrics in three tasks, and even outperforms GPT-4 to generate more complex and abstract rules.

\subsection{Downstream Reasoning Evaluation}
We further analyze the effectiveness of our inference engine in generating logical rules or explanations to enhance downstream reasoning tasks. We evaluate on following commonsense reasoning datasets: StrategyQA~\cite{geva2021did}, SOCIAL IQA~\cite{sap2019socialiqa}, LINK~\cite{li2023search}, PIQA~\cite{bisk2020piqa} and CSQA2.0~\cite{talmor2022commonsenseqa}. 
We use a zero-shot CoT strategy to prompt two baseline models, Mistral-7B-Instruct-v0.1 and Llama-2-7b-chat~\cite{touvron2023llama}, to answer questions with following explanations. We then utilize our inference engine to generate logical rules or explanations relevant to answer questions, and supplement these generated rationals to baseline models as input to enhance their performance. We compare the prediction accuracy of our inference engine augmented models against baselines. 

The comparative results are shown in Tabel~\ref{downstream_tasks}. Our inference engine can generate logical rules or explanations that benefit multiple downstream commonsense reasoning tasks on top of different backbone models. 
For the lack of clear advantage on PIQA and performance decline on CSQA2.0, we speculate that PIQA may be contaminated during Mistral's training process, and CSQA2.0's focus is mainly on longtail commonsense knowledge rather than requiring logical rules inference, like "Is cotton candy sometimes made out of cotton?" 
\begin{table}[th!]
    \centering
    \setlength{\tabcolsep}{3pt}
    \resizebox{0.48\textwidth}{!}{
    \begin{tabular}{c|cc|cc}
        \toprule
        \multirow{2}{*}{Dataset} & Mistral & Mistral+rules & LLama & LLama+rules \\
        & \multicolumn{2}{c|}{(Mistral-7b)} & \multicolumn{2}{c}{(LLama2-7b)}  \\
         \midrule
        StrategyQA & 54.50 & \bf 56.75 & 58.00 & \bf 60.48 \\
        SOCIAL IQA & 64.00 & \bf 68.50 & 53.50 & \bf 60.50 \\
        LINK head & 53.68 & \bf 68.38 & 58.09 & \bf 70.59 \\
        LINK longtail & 53.33 & \bf 67.50 & 55.83 & \bf 65.00 \\
        \midrule
        PIQA & 65.00 & 65.00 & 58.5 & \bf 62.0 \\
        CSQA2.0 & 59.00 & \bf 62.50 & 64.00 & 60.00 \\
        \bottomrule
    \end{tabular}
     }
    \caption{Downstream reasoning performance.}
    \label{downstream_tasks}
\end{table}

 
\section{Related Work}
\noindent\textbf{Logical Rule Generation} Logical inferential rules are crucial for everyday reasoning~\cite{geva2021did, talmor2022commonsenseqa}, and collecting these inferential rules is challenging. Prior work mainly adopts inductive logic programming (ILP)~\cite{yang2019learn, qu2020rnnlogic, sen2022neuro} for rule generation. However, they can only generate rules from existing knowledge graphs and the generated rules has potential inaccuracies.
Alternatively, \citet{sinha2019clutrr} manually create a set of inferential rules for inductive reasoning, but their scope is limited to kinship. \citet{sap2019atomic} construct a commonsense inferential rule base through crowdsourcing, but these rules tend to be overly simple and specific, struggling to move beyond basic intuition and generalize to varied situations. Abstract and complex rules are essential in tackling diverse complex questions, paving the way for complex reasoning and decision-making. Although LLMs have opened new avenues for generating inferential rules~\cite{zhu2023large}, they still struggle to automatically craft abstract and complex rules. 

\vspace{1mm}
\noindent\textbf{Integration of Logical Rules and LLMs} Recently, the integration of inferential rules with neural networks, particularly LLMs, has gained significant attention. This approach combines the logical interpretability of symbolic reasoning and adaptive power of neural computing, improving LLMs' logical reasoning ability~\cite{wang2022lsat, mu2023can}. ~\citet{wang2021logic} and \citet{olausson2023linc} transform textual statements into logical expressions and conduct symbolic reasoning following logical rules. \citet{weir2022dynamic} train neural models using a set of inferential rules for dynamic application. This direction broadens LLMs' ability with flexible rule generation and application for complex reasoning.

\section{Conclusion}
This paper examines the proficiency of GPT-series LLMs in capturing logical inferential rules and probes their challenging reasoning space. We introduce a logic scaffolding inferential rule generation (LOIRE) framework to create an inferential rule base ULogic, including nearly 8,000 primitive rules and over 6,000 compositional rules across five domains. 
Our evaluations using a subset of ULogic show that even advanced models like GPT-4 struggle with compositional and structural complex rules and exhibit certain biases.
Furthermore, we distill ULogic into a smaller inference engine that performs well in generating inferential rules and benefit downstream reasoning tasks. Our work points out where LLMs need to improve in logical reasoning and offers a pathway to enhance their reasoning capabilities.

\section*{Limitations}

\paragraph{Limitation on inferential rule coverage.}
Commonsense inferential rules may exist in diverse formats and span various domains. Our work mainly focuses on rules formatted as \textit{if-then} statements, covering five domains: object affordance, accessibility, interaction, location and person's need.
In future work, we will expand our scope to include inferential rules of other formats and explore additional domains for broader coverage.

\paragraph{Limitation on probing open-source models.}
Our work does not probe and analyze open-source models. While GPT-4 and GPT-3.5-turbo are considered as the most advanced models, open-source counterparts may exhibit different behaviors or patterns in understanding inferential rules with varying complexities. These aspects will be the subject of future exploration. 

\paragraph{Risk of environmental impact}
A significant risk associated with our framework and analysis is the potential increase in environmental burdens due to the extensive use of OpenAI's APIs for LLMs. This impact can be mitigated by replacing GPT-4 with future smaller-scale open-source models that are more efficient with less environmental impact.

\paragraph{Potential error in rule generation.}
Generating inferential rules with specific requirements poses a significant challenge. As the majority of our framework's pipeline are powered by GPT-4, it may inevitably generate inferential rules with logical inaccuracies even incorporating human verification. This might result in less accurate probing of LLMs.

\section*{Ethical Consideration}
All rules we collected through LLMs are released publicly for usage and its probing subset for proficiency analysis have been subjected to a thorough review by the authors. The code of our generation pipeline and probing experiments will also be publicly released.
This setting guarantees transparency and reproducibility in our experiments, allowing other researchers to evaluate and expand upon our work.
Our logic scaffolding framework is strictly limited to be used for rule generation that follow the ethical guidelines of the community. The authors emphatically denounce the use of our framework for generating inaccurate or harmful rules.

\section*{Acknowledgments}
Siyuan Wang's research is supported by National Natural Science Foundation of China (No. 62176058) and National Key R\&D Program of China (2023YFF1204800). The project's computational resources are supported by CFFF platform of Fudan University. 
Xiang Ren's research is supported in part by the Office of the Director of National Intelligence (ODNI), Intelligence Advanced Research Projects Activity (IARPA), via the HIATUS Program contract \#2022-22072200006, the Defense Advanced Research Projects Agency with award HR00112220046, and NSF IIS 2048211.

\bibliography{anthology,custom}

\newpage
\appendix

\section{Primitive Rule Generation Pipeline}
\subsection{Abstract Objects and Common Properties}
\label{appen:abstract_objects}
Table~\ref{table_abstract_objects} list 32 most common abstract objects and 18 common properties for primitive rule generation.
\begin{table}[!h]
    \begin{center}
    \resizebox{0.49\textwidth}{!}{
    \begin{tabular}{m{1.58cm}m{8cm}}
    \toprule 
    \bf Type & \bf Words \\
    \midrule
    Abstract Objects & ``Person'', ``Animal'', ``Plant'', ``Food'', ``Alcohol'', ``Disease'', ``Drug'', ``Natural Phenomenon'', ``Condition'', ``Material'', ``Substance'', ``Furniture'', ``Publication'', ``Organization'', ``Authorization'', ``Facility'', ``Natural Place'', ``Event'', ``Show'', ``Artwork'', ``Job'', ``Game'', ``Vehicle'', ``Tool'', ``Technology'', ``Electronic Device'', ``Platform'', ``Financial Product'', ``Skill'', ``Legislation'', ``Region'', ``Time Period'' \\
    \midrule
    Common Properties & ``Age'', ``Price'', ``Money'', ``Height'', ``Length'', ``Weight'', ``Strength'', ``Size'', ``Density'', ``Volume'', 
    ``Temperature'', ``Hardness'', ``Speed'', ``BoilingPoint'', ``MeltingPoint'', ``Frequency'', ``Decibel'', ``Space'' \\
    \bottomrule
    \end{tabular}
    }
\caption{\label{table_abstract_objects} List of pre-defined abstract objects and common properties.}
    \end{center}
\end{table}

\subsection{Rule Domains}
\label{rule_domain}
Table~\ref{table_five_domains} illustrates the detailed explanations, example predicates and rules across five domains.
\begin{table*}
    \begin{center}
    \resizebox{0.96\textwidth}{!}{
    \small 
    \begin{tabular}{m{1.5cm}m{3.7cm}m{4.3cm}m{6cm}}
    \toprule 
    \bf Domain & \bf Explanation & \bf Predicates & \bf Examples \\
    \midrule
    Object \newline  Affordance & Whether a person can take an action over an object based on its property and requirement & CanDrive(Person X, Vehicle Y); \newline CanCreate(Person X, Artwork Y); \newline CanAttend(Person X, Event Y); & 
    CanDrive(Person X, Vehicle Y):- Have(Person X, Age Z1), RequireMinimumAge(Vehicle Y, Age Z2), BiggerThan(Age Z1, Age Z2);\\
    \midrule
    Object \newline Accessibility & Whether an object can access the other object based on its physical condition, spatial and temporal restriction & CanAccess(Person X, Show Y); \newline CanAccess(Animal X, Tool Y); \newline CanAccess(Animal X, Animal Y); & CanAccess(Person X, Show Y):- LocatedIn(Person X, Region Z), BroadcastIn(Show Y, Region Z); \quad CanNotAccess(Person X, Tool Y):- AllergicTo(Person X, Material Z), MadeOf(Tool Y, Material Z); \\
    \midrule
    Object \newline Interaction & How an object can interact with 
    the other object based on their physical, spatial or temporal properties & CanSubmergeIn(Substance X, \newline Substance Y);  \quad CanAdaptedFrom(Show X, Artwork Y); \newline CanFitIn(Tool X, Tool Y); & CanSubmergeIn(Substance X, Substance Y):- DensityOf(Substance X, Density Z1), DensityOf(Substance Y, Density Z2), BiggerThan(Density Z1, Density Z2);
    \\
    \midrule
    Object \newline Location & The location description of an object & OriginatedFrom(Food X, Region Y); BannedIn(Drug X, Region Y); \newline
    BornIn(Person X, Region Y); & OriginatedFrom(Food X, Region Y):- ProcessedIn(Food X, Facility Z), LocatedIn(Facility Z, Region Y); \\
    \midrule
    Person's \newline Need & Person need to take an action over objects under a specific circumstance & NeedToConsume(Person X, Drug Y); \newline NeedToWater(Person X, Plant Y); & NeedToConsume(Person X, Drug Y):- Has(Person X, Disease Z), CanTreat(Drug Y, Disease Z); \\
    \bottomrule
    \end{tabular}
    }
\caption{\label{table_five_domains} The explanations, example predicates and rules of five different domains.}
    \end{center}
\end{table*}

\subsection{Prompts for Premise Generation}
\label{rule_generation_prompts}
For premise generation in each domain, we design an instruction followed by two demonstrations to iteratively prompt GPT-4, and the underlined sentence is the rule description which varies according to the specific domain, as shown in Table~\ref{domain_generation_prompts}.

\begin{table*}[h!]
\centering
\resizebox{0.98\textwidth}{!}{
\begin{tcolorbox}[colback=blue!5!white,colframe=black,width=0.98\textwidth,title={Instruction for Premise Generation (Object Affordance)}]
\small
According to commonsense knowledge in realistic scenarios, please generate 2 logical rules in both Prolog and natural langauge to describe the premises of the given conclusion. The rules in Prolog should have the same meaning with the rules in natural language. \\
Each rule should contain multiple premises and each premise should contain two variables in (X, Y, Z, Z1, Z2). \\
\underline{The rules should describe object affordance based on its property (such as height, age, price) and requirement (such} \underline{as required skill, source, tool).} \\
The premises should not contain negative words such as 'not', 'no', 'never' and 'un-' \\

Conclusion: \{conclusion\} \\
Rules:
\end{tcolorbox}
}
\vspace{3mm}

\centering
\resizebox{0.98\textwidth}{!}{
\begin{tcolorbox}[colback=blue!5!white,colframe=black,width=0.98\textwidth,title={Demonstrations for Premise Generation (Object Affordance)}]
\small
Conclusion: CanCook(Person X, Food Y) \\
Rules: \\
1. CanCook(Person X, Food Y):- CanUse(Person X, Tool Z), UsedForCook(Tool Z, Food Y); \\
If Person X can use Tool Z which is used for cooking Food Y, then Person X can cook Food Y. \\
2. CanCook(Person X, Food Y):- Master(Person X, Skill Z), RequiredForCooking(Skill Z, Food Y); \\
If Person X has mastered Skill Z which is required for cooking Food Y, then Person X can cook Food Y. \\
\\
Conclusion: CanDrive(Person X, Vehicle Y) \\
Rules: \\
1. CanDrive(Person X, Vehicle Y):- Have(Person X, Age Z1), RequireMinimumAge(Vehicle Y, Age Z2), BiggerThan(Age Z1, Age Z2); \\
If Person X has Age Z1 and the minimum age requirement for driving Vehicle Y is Age Z2, Age Z1 is bigger than Age Z2, then Person X can drive Vehicle Y. \\
2. CanDrive(Person X, Vehicle Y):- Obtain(Person X, Authorization Z), RequiredForDriving(Authorization Z, Vehicle Y); \\
If Person X have obtained a specific Authorization Z and Authorization Z is required for driving Vehicle Y, then Person X can drive Vehicle Y. 
\end{tcolorbox}
}

\vspace{4mm}
\resizebox{1.0\textwidth}{!}{
\small 
\begin{tabular}{m{2.55cm}m{13.5cm}}
    \toprule
    \bf Domain & \bf Rule Description \\
    \midrule
    Object Affordance & The rules should describe object affordance based on its property (such as height, age, price) and requirement (such as required skill, source, tool). \\
    Object Accessibility & The rules should describe object accessibility based on its physical condition, spatial and temporal restriction. \\
    Object Interaction & The rules should describe object interaction based on its physical, spatial or temporal properties (such as speed, hardness, density, height, time period). \\
    Object Location & The rules should describe the location information of an object. \\
    Person's Need & The rules should describe person's need to take an action over the object. \\
\bottomrule
\end{tabular}
}
\caption{Prompts for rule generation in different domains.}
\label{domain_generation_prompts}
\vspace{-2mm}
\end{table*}

\subsection{Grammatical Validity for Rule Filtering}
\begin{figure}[h!]
    \centering
    \includegraphics[width=0.65\columnwidth]{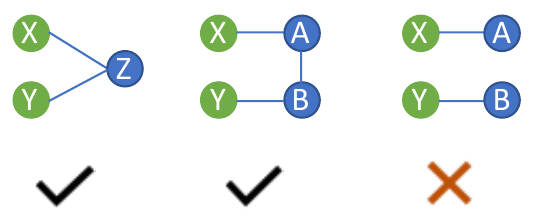}
    \caption{Grammatically valid and invalid rule graphs.}
    \label{valid_rules}
\end{figure}

As Figure~\ref{valid_rules}, we check whether the variables in premises form a connected graph from node ``X'' to node ``Y'' to filter grammatically invalid rules.
\label{appen:grammar_validity}

\subsection{Prompts for Rule Filtering}
\label{rule_filtering_prompts}
Table~\ref{prompt_rule_filtering} is an example prompt for rule filtering in object affordance domain.
\begin{table*}[h!]
\centering
\resizebox{0.98\textwidth}{!}{
\begin{tcolorbox}[colback=blue!5!white,colframe=black!55!black,width=0.98\textwidth,title={Prompt for Rule Filtering}]
\small
True or False? Please predict whether the input rule is accurate or not according to commonsense knowledge in realistic scenarios, and also explain why. \\
Examples: \\
Input: If Person X has an Age Z1 and Vehicle Y requires an Age above Z2 for driving, with ... \\
Output: True. Because Person X has achieved the ... \\
Input: If Person X was born in Season Z and Plant Y blooms in the same Season Z, then Person X can access Plant Y. \\
Output: False. Because a person's birth season and a plant's blooming season has no logical connection. \\

Input: \{candidate rule\} \\
Output:
\end{tcolorbox}
}
\caption{A prompt for rule filtering in object affordance.}
\label{prompt_rule_filtering}
\end{table*}

\begin{figure*}[!t]
    \centering
    \includegraphics[width=1.0\textwidth]{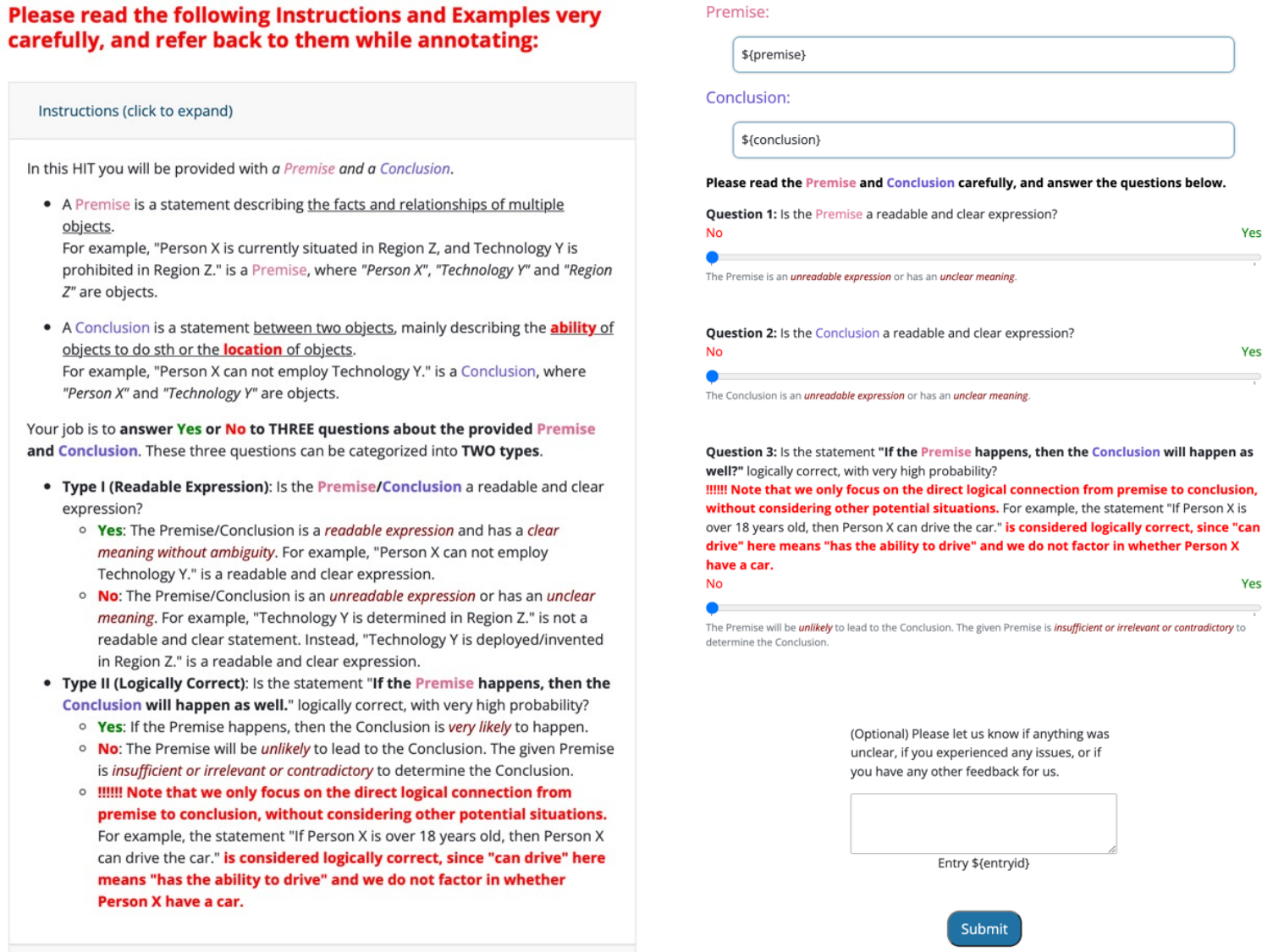}
    \caption{AMT template for human verification of primitive rules.}
    \label{human_verify_figure}
\end{figure*}
\subsection{Human Verification Templates and Rates}
\label{human_verify_template}
Before human verification, we first craft a qualification task to select AMT annotators from all English-speaking countries (US, UK, New Zealand, Australia, Canada). The prospective workers are presented with three representative test cases and need to predict whether the premise and conclusion are clearly readable, and if the premise logically entails the conclusion. Only those workers correctly passing all the test cases are recruited.  We compensated workers with different rates based on the complexity of the annotated rules, with prices ranging from 0.08 to 0.16 dollars (i.e., 0.08, 0.10, 0.12, 0.14, 0.16) for rules with complexity levels from 1 to 5. On average, this amounts to 0.12 dollars per annotation. Each annotation takes approximately 0.5 minutes to complete, aiming to match a rate of 15 dollars per hour based on their working time. 

The detailed template for human verification is shown as Figure~\ref{human_verify_figure}. 
This template is also used for getting human performance in rule probing analysis, wherein a separate cohort of workers is qualified for manual rule probing. Besides, the overall rates of rule acceptance in different domains during human verification are listed Table~\ref{yield_rule}.
\begin{table*}[h!]
    \centering
    \setlength{\tabcolsep}{3pt}
    \resizebox{0.7\textwidth}{!}{
    \begin{tabular}{c|ccccc}
        \toprule
         & Affordance & Accessibility & Interaction & Location & Person's Need \\
         \midrule
        Yield Rate & 48.09 & 37.28 & 52.81 & 53.74 & 49.45 \\
        \bottomrule
    \end{tabular}
     }
    \caption{The rule yield rates (\%) of human verification.}
    \label{yield_rule}
\end{table*}

\subsection{Ablation on Rule Filtering}
We calculated the rule yield rates for both heuristic filtering and self-critic, which are respectively 77.59\% and 80.39\%. Together, they totally yield 62.38\% of candidate generated rules through the whole rule filtering process.
We sample 100 rules before the self-critic and obtain 79 rules after the self-critic, and conduct human verification over them. Before the self-critic, 42 out of 100 rules (42\%) are valid. After the self-critic, 40 out of 79 rules (50.63\%) are valid. Among the 21 filtered rules, our self-critic can effectively identify 19 invalid rules, thereby enhancing the quality of retained rules. 

\section{Rule Probing}
\subsection{Rule Probing Templates}
\label{appen:probing_templates}
Table~\ref{table_probing_templates} lists five different templates for unbiased rule probing.
\begin{table}[!h]
    \begin{center}
    \resizebox{0.49\textwidth}{!}{
    \begin{tabular}{m{0.3cm}m{6.7cm}cp{1.2cm}}
    \toprule 
    & \bf Template & \bf Label \\
    \midrule
    1 & True or False? Please predict whether the input rule is very likely to be true. & True/False \\
    2 & Right or Wrong? Please predict whether the input rule is valid and correct. & Right/Wrong \\
    3 & Yes or No? Please predict whether the premise entails the conclusion. & Yes/No \\
    4 & Premise:..., Conclusion:... Does premise entail conclusion? Please answer Yes or No. & Yes/No \\
    5 & Given the observations ..., can we draw the conclusion ...? Please answer Yes or No. & Yes/No \\
    \bottomrule
    \end{tabular}
    }
\caption{\label{table_probing_templates} Five templates for rule probing.}
    \end{center}
\end{table}

\subsection{Dual-side Rule Probing Setting}
\label{appen:both_side_probing}
Table~\ref{table:both_side_probing} illustrate a concrete example of dual-side rule probing.
\begin{table}[!h]
    \resizebox{0.49\textwidth}{!}{
    \begin{tabular}{m{5.8cm}cp{1cm}}
    \toprule 
    \emph{If Premise, then Conclusion$\_$original.} & True/Right/Yes \\
    \emph{If Premise, then Conclusion$\_$flipped.} & False/Wrong/No \\
    \midrule
   
    \multicolumn{2}{c}{\bf Example} \\
    \midrule
    \emph{If Person X is allergic to Substance Z and Food Y contains Substance Z, then Person X \textbf{cannot eat} Food Y.} & True/Right/Yes \\
    \midrule
    \emph{If Person X is allergic to Substance Z and Food Y contains Substance Z, then Person X \textbf{can eat} Food Y.} & False/Wrong/No \\
    \bottomrule
    \end{tabular}
    }
    \caption{A specific example of dual-side rule probing.}
    \label{table:both_side_probing}
\end{table}


\subsection{Rule Depths Probing}
\label{appen:rule_depth_probing}
\begin{figure}[h!]
    \centering
    \includegraphics[width=0.78\columnwidth]{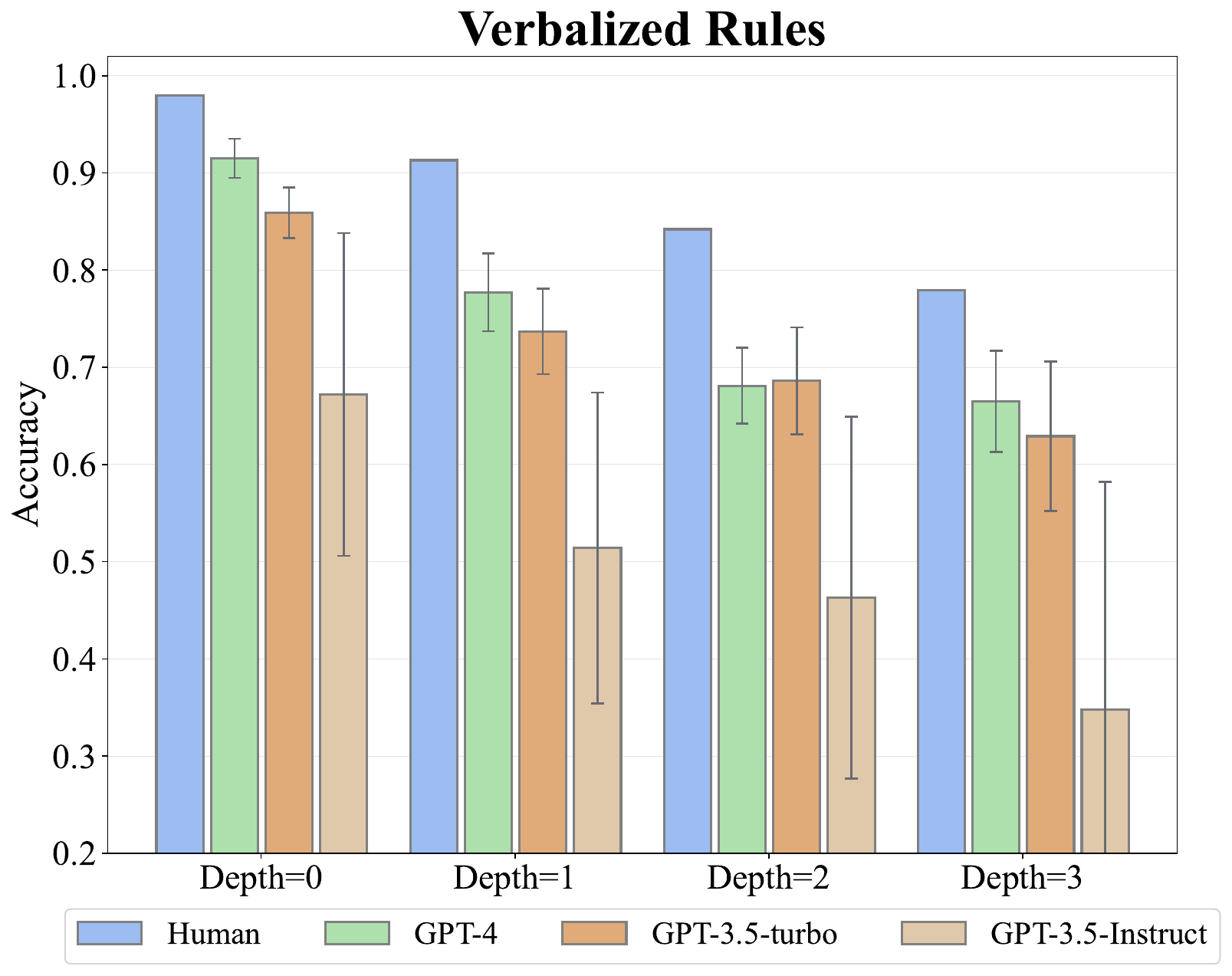}
    \caption{Probing results of varied depths.}
    \label{depth_analysis}
\end{figure}

The analysis of GPT-series LLMs and human on different compostional depths is presented as Figure~\ref{depth_analysis}.  ``Depth=0'' represents primitive rules and ``Depth=1,2,3'' denote compositional rules involving 1 to 3 backward chaining steps. 

\subsection{Illustrations of Rule Structures}
\label{appen:rule_structure}
\begin{figure*}[t!]
    \centering
    \includegraphics[width=1.0\textwidth]{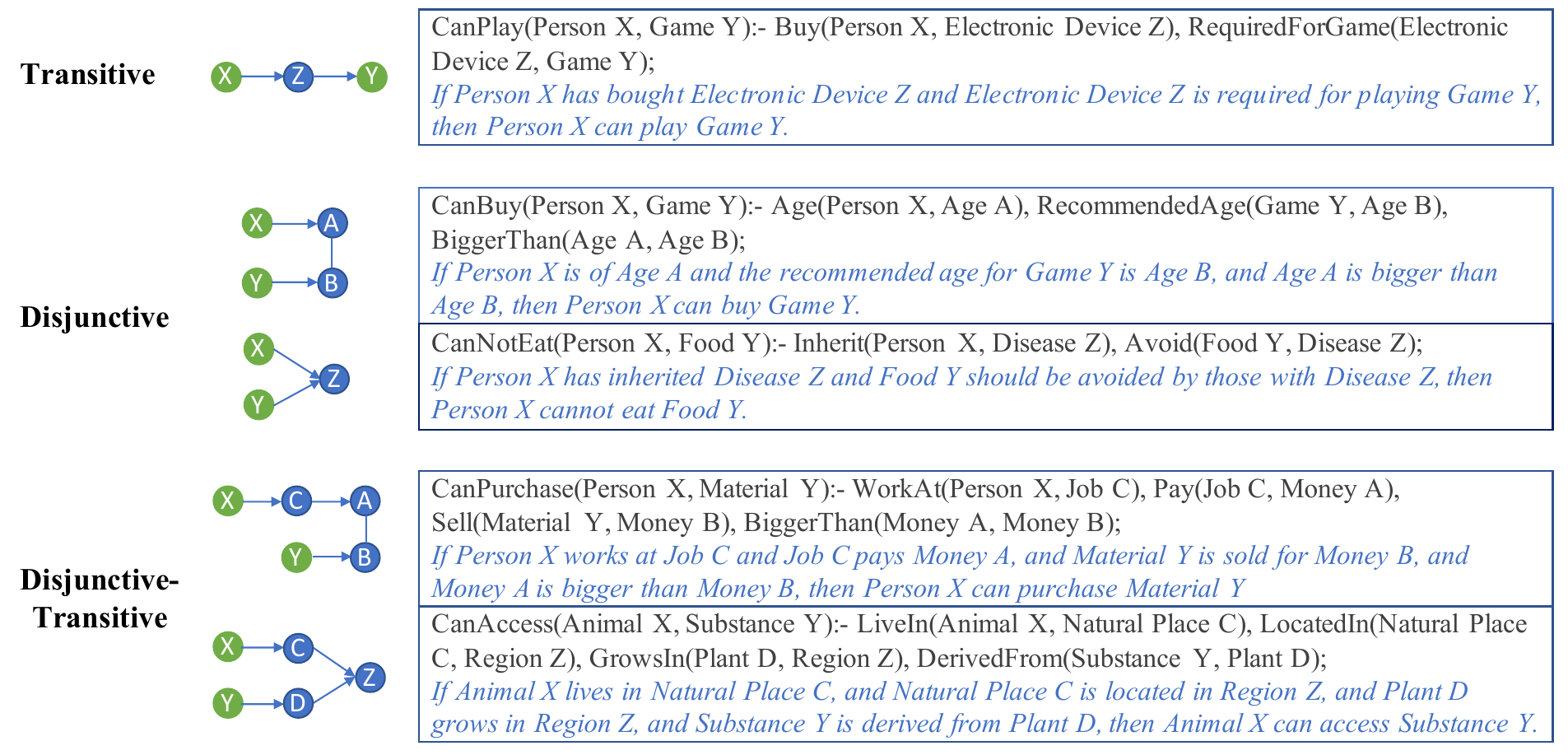}
    \caption{Example rules of different structures.}
    \label{rule_structure_figure}
\end{figure*}

Figure~\ref{rule_structure_figure} displays several examples showcasing both symbolic and verbalized rules across different structure types.

\subsection{Different CoT Prompts}
\label{appen_cot_prompts}
Table~\ref{cot_prompts} lists different prompts of three CoT strategies for rule probing.
\begin{table}[h!]
    \centering
    \resizebox{0.48\textwidth}{!}{
    \begin{tabular}{m{2.1cm}m{7.55cm}}
    \toprule 
    \bf CoT strategy & \bf Prompt \\
    \midrule
    Answer-Explain & True or False? Please predict whether the input rule is very likely to be true, and also explain why. \\
    \midrule
    Answer-Explain  & True or False? Please predict whether the input rule is very likely to be true. Please first briefly explain your thought process in one sentence, and then give your answer. \\
    \midrule
    Self-Consistency & True or False? Please predict whether the input rule is very likely to be true. Please first generate three different sentences to respectively explain your three thought processes briefly, and then based on the corresponding thought to give your answer. Finally, output the final answer according to majority voting. \\
    \bottomrule
    \end{tabular}
    }
    \caption{Three CoT prompts for rule probing.}
    \label{cot_prompts}
\end{table}

\subsection{Necessary Bias}
\label{necessary_bias}
As mentioned in Section~\ref{sec3_2_analysis}, We investigate why GPT-4 significantly underperforms GPT-3.5-Turbo on transitive rules. Transitive rules typically describe a straightforward logical chain from variable X to Y, where GPT-3.5-Turbo and GPT-3.5-Instruct manage with greater ease. 
In contrast, we find that GPT-4 exhibits a ``necessary bias'' that tend to consider all necessary conditions reaching a conclusion, which avoids drawing a definite judgement. This conservative response is more pronounced in transitive rules, where GPT-4 more frequently responds with hesitations like "it does not necessarily mean" in its explanations. We present a probing example of a transitive rule by GPT-4 in Table~\ref{appendb_3}, including its generated prediction, explanation and the corresponding label. We hypothesize that this conservative style may come from LLMs' preference alignment during Reinforcement Learning with Human Feedback~\cite{ouyang2022training}.

\begin{table}[!h]
    \resizebox{0.49\textwidth}{!}{
    \begin{tabular}{m{1.8cm}m{7.6cm}}
    \toprule 
    Input & True or False? Please predict whether the input rule is very likely to be true, and also explain why. \newline
    Rule: If Person X has a job A that requires Skill Z, and Skill Z is required to cook Food Y, then Person X can cook Food Y. \\
    \midrule
    Prediction & False \\
    Explanation & Just because Person X has a job that requires Skill Z, \underline{it does not necessarily mean} they can cook Food Y. They may have the skill, but cooking also requires other knowledge of recipes, techniques. \\
    \midrule
    Label & True \\
    \bottomrule
    \end{tabular}
    }
    \caption{An example probing result of a transitive rules by GPT-4.}
    \label{appendb_3}
\end{table}

\begin{table*}[ht!]
\begin{tcolorbox}[colback=blue!5!white,colframe=black,width=1.0\textwidth,title={Prompt for Conclusion Generation}]
\small
Given the premise, please generate its conclusion between X and Y in both Prolog and natural language. \\
The conclusion in Prolog should have the same meaning with the conclusion in natural language. \\
Each conclusion should contain only two variables X and Y without mentioning other variables, like A, B, C, Z. \\

\#\#\# Examples: \\
Premise: If Person X is allergic to Material Z and Furniture Y is made from Material Z. \\
Conclusion: \\
\text{[}Prolog\text{]}: CanNotHold(Person X, Furniture Y); \\
\text{[}Natural Language\text{]}: Person X cannot hold Furniture Y. \vspace{1.5mm}\\
Premise: If Substance X has a Density Z1, the density of Substance Y is Density Z2, and Density Z1 is bigger than Density Z2. \\
Conclusion: \\
\text{[}Prolog\text{]}: CanSubmerge(Substance X, Substance Y); \\
\text{[}Natural Language\text{]}: Substance X can submerge in Substance Y. \\
\newline
\\
Premise: \{premise\} \\
Conclusion:
\end{tcolorbox}
\caption{Prompt ChatGPT and GPT-4 for conclusion generation.}
\label{conc_gen_prompt}
\end{table*}
\section{Inference Engine}
\subsection{Illustration of Instruction Tuning}
\label{appen:instruction_tuning_pipeline}
\begin{figure}[th!]
    \centering
    \includegraphics[width=0.78\columnwidth]{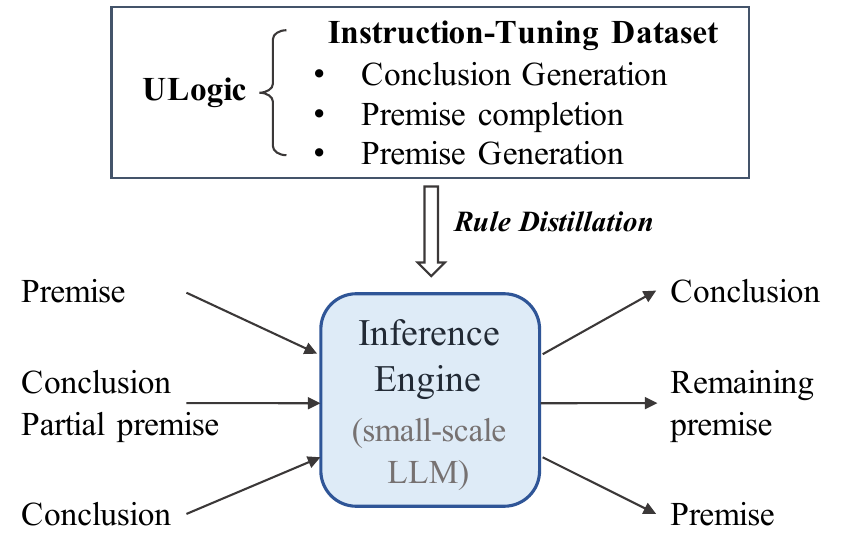}
    \caption{Rule distillation for inference engine.}
    \label{instruction_tuning_pipeline}
\end{figure}

\begin{figure*}[!h]
    \centering
    \includegraphics[width=0.68\textwidth]{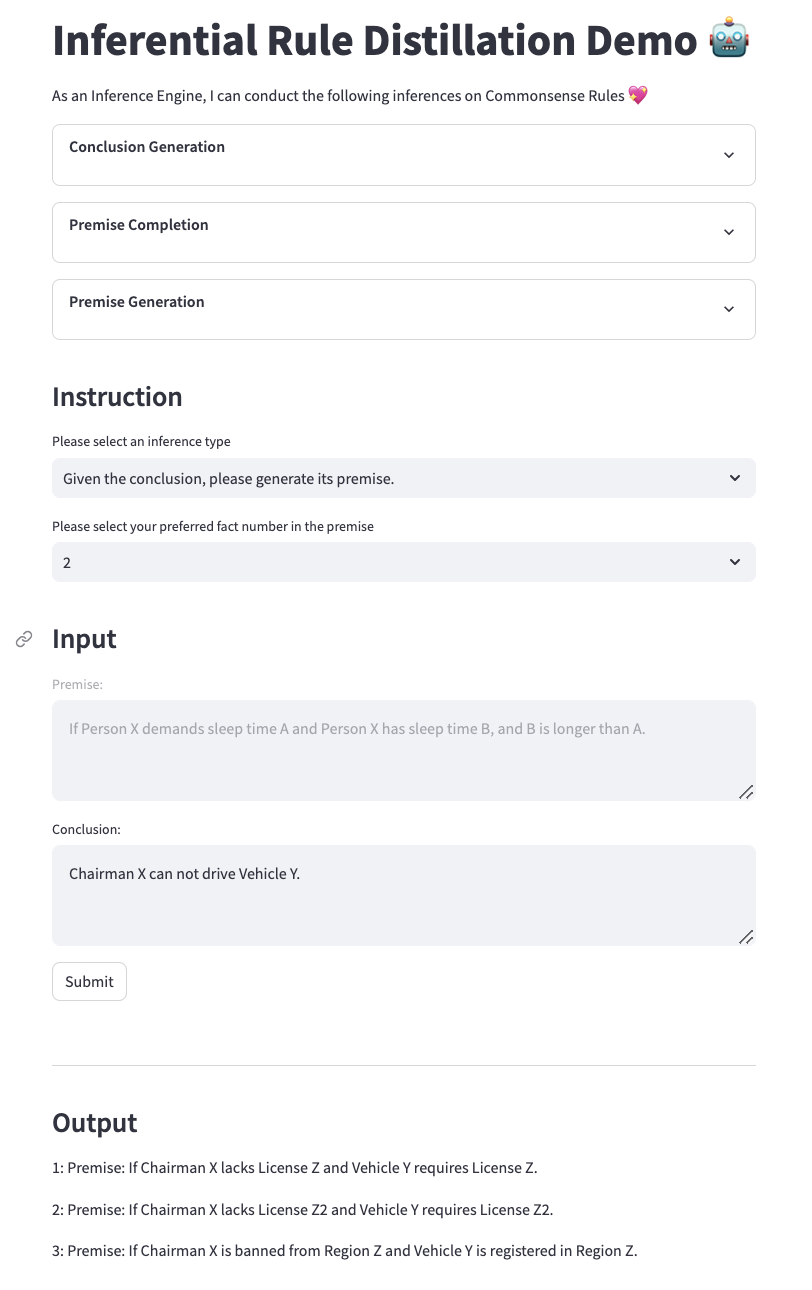}
    \caption{Inference Engine Demo.}
    \label{inference_engine_demo}
\end{figure*}

Figure~\ref{instruction_tuning_pipeline} illustrate the pipeline of instruction tuning for rule distillation as an inference engine. Our inference engine is trained for three tasks: conclusion generation, premise completion and premise generation. The conclusion generation focuses on creating a conclusion from a provided premise. For premise completion, given a conclusion and its partial premise, the inference engine must complete the remaining premise part to support the conclusion. In premise generation, the engine is tasked with creating premises of varying complexity based on a given conclusion, specifically generating premises with one, two or even more facts. We also provide an inference engine demo for flexible rule generation as shown in Figure~\ref{inference_engine_demo}.

\subsection{Implementation Details}
\label{appen:implementation_details}
We fine-tune Mistral-7b with our constructed instruction dataset with Quantization LoRA (QLoRA) method~\cite{hu2021lora,dettmers2023qlora} as our inference engine. We set the learning rate to $7\times10^{-5}$, batch size to 8, gradient accumulation step to 16, and train the model 2 epochs. We apply QLoRA to all the linear layers of the model, including q$\_$proj, k$\_$proj, v$\_$proj, o$\_$proj, gate$\_$proj, up$\_$proj, down$\_$proj, and lm$\_$head. The $\alpha$ and $r$ of the QLoRA method are both set to 16.

\subsection{Prompting ChatGPT and GPT-4 for Three Tasks}
\label{three_task_prompts}
As Step-2 of Sec.~\ref{chap_rule_generation}, we utilize two-shot prompts to instruct ChatGPT and GPT-4 in simultaneously generating symbolic and verbalized responses for three tasks, as shown in Table~\ref{conc_gen_prompt},~\ref{prem_comp_prompt},~\ref{prem_gen_prompt}.

\begin{table*}
\begin{tcolorbox}[colback=blue!5!white,colframe=black,width=1.0\textwidth,title={Prompt for Premise Completion}]
\small
Given the conclusion and a part of its premise, please complete the remaining portion of the premise in both Prolog and natural language. \\
The remaining premise in Prolog should have the same meaning with the remaining premise in natural language. \\
Each fact in the remaining premise should contain two variables, like X, Y, Z, Z1, Z2, A, B. \\

\#\#\# Examples: \\
Conclusion: Person X cannot use Furniture Y. \\
Partial Premise: If Person X is allergic to Material Z, \\
Remaining Premise: \\
\text{[}Prolog\text{]}: MadeFrom(Furniture Y, Material Z); \\
\text{[}Natural Language\text{]}: Furniture Y is made from Material Z. \\

Conclusion: Substance X can submerge in Substance Y. \\
Partial Premise: If Substance X has a Density Z1, the density of Substance Y is Density Z2, \\
Remaining Premise: \\
\text{[}Prolog\text{]}: BiggerThan(Density Z1, Density Z2); \\
\text{[}Natural Language\text{]}: Density Z1 is bigger than Density Z2. \\
\newline
\\
Conclusion: \{conclusion\} \\
Partial Premise: \{partial premise\} \\
Remaining Premise: 
\end{tcolorbox}
\caption{Prompt ChatGPT and GPT-4 for premise completion.}
\label{prem_comp_prompt}
\end{table*}

\begin{table*}[h!]
\begin{tcolorbox}[colback=blue!5!white,colframe=black,width=1.0\textwidth,title={Prompt for Premise Generation}]
\small
Given the conclusion, please generate three different premises in both Prolog and natural language, ensuring that each Prolog premise conveys the same meaning as its natural language counterpart. \\
Each premise should contain a specified number of facts, with each fact comprising only two variables, such as X, Y, Z, Z1, Z2, A, B. \\

\#\#\# Examples: \\
Fact number: 1 fact \\
Conclusion: Person X has Skill Y. \\
Three Premises: \\
1. [Prolog] Learned(Person X, Skill Y); [Natural Language] If Person X learned Skill Y. \\
2. [Prolog] Inherit(Person X, Skill Y); [Natural Language] If Person X inherits Skill Y. \\
3. [Prolog] Acquire(Person X, Skill Y); [Natural Language] If Person X acquires Skill Y. \\

Fact number: more than 2 facts \\
Conclusion: Person X cannot attend Event Y. \\
Three Premises:  \\
1. [Prolog] Have(Person X, Age Z1), RequireMinimumAge(Event Y, Age Z2), BiggerThan(Age Z2, Age Z1); [Natural Language] If Person X has Age Z1 and the minimum age requirement for attending Event Y is Age Z2, Age Z2 is bigger than Age Z1. \\
2. [Prolog] Have(Person X, Height Z1), RequireAbove(Event Y, Height Z2), SmallerThan(Height Z1, Height Z2); [Natural Language] If Person X has a Height Z1, and Event Y requires a Height above Z2, and Height Z1 is smaller than Height Z2. \\
3. [Prolog] HaveCriminalRecord(Person X, Event Z), ProhibitedBy(Event Z, Legislation A), EnforcedIn(Legislation A, Region B), HeldIn(Event Y, Region B); [Natural Language] If Person X has a criminal record for Event Z and Event Z is prohibited by Legislation A, which is enforced in Region B, and Event Y is held in Region B. \\
\newline
\\
Fact number: \{fact num\} \\
Conclusion: \{conclusion\} \\
Three Premises: 
\end{tcolorbox}
\caption{Prompt ChatGPT and GPT-4 for premise generation.}
\label{prem_gen_prompt}
\end{table*}

\subsection{Evaluation Metrics}
\label{appen:evaluation_metrics}
We detailed describe the metrics for evaluating our inference engine against ChatGPT and GPT-4 for the premise generation task.
\begin{itemize}[itemsep=0.2pt, leftmargin=10pt, parsep=2pt, topsep=2pt]
    \item Accuracy: The premise is logically correct to infer the conclusion and follow the instruction regarding the specific number of facts.
    \item Diversity: The degree of variation among the three generated rules.
    \item Complexity: Assessed only for premise generation with more than 2 facts, measuring the fact number and the semantic difficulty.
    \item Abstractness: The variable types in premises are abstract to generalize to diverse instances. For example, the variable types `` Region'' and ``Event'' are abstrct while ``New York'' and ``The FIFA World Cup'' are specific entities with low abstractness.
\end{itemize}

\subsection{LLM Evaluation Prompts}
\label{llm_evaluation_appen}
We prompt GPT-4 as the evaluator for rating the accuracy of conclusion generation and premise completion tasks, and the accuracy, diversity, complexity and abstractness of the premise generation task. We adopt one-shot prompts which are shown as Table~\ref{accuracy_rating} and Table~\ref{diversity_complexity_rating} (with demonstrations omitted).
\begin{table*}[h!]
\begin{tcolorbox}[colback=blue!5!white,colframe=black,width=1.0\textwidth,title={Prompt for Rating the Accuracy of Conclusion Generation}]
\small
You are a helpful scoring assistant. \\
Please read the provided premise carefully, and rate the accuracy of the candidate conclusion on a scale of 1 to 3: \\
- 1 (not accurate): The conclusion is clearly unsupported, irrelevant or contradictory to the provided premise. \\
- 2 (somewhat accurate): The conclusion, despite being supported by the premise, fails to state the definitive link between X and Y, or contradicts common sense, or lacks clarity. \\
- 3 (highly accurate): The conclusion correctly states the definitive link between X and Y, and is well-supported by the premise aligning with both established facts and common sense. \\
Please first output your rating based on your general knowledge and logical reasoning, and then provide a brief explaination with no more than 100 words. \\

\text{[}Provided Premise\text{]}: \{premise\} \\
\text{[}Candidate Conclusion\text{]}: \{conclusion\} \\
\text{[}Output\text{]}:
\end{tcolorbox}

\begin{tcolorbox}[colback=blue!5!white,colframe=black,width=1.0\textwidth,title={Prompt for Rating the Accuracy of Premise Completion}]
\small
You are a helpful scoring assistant. \\
Please read the provided conclusion and its partial premise carefully, and rate the accuracy of its remaining premise in completing the provided premise to reach the conclusion, using a scale from 1 to 3: \\
- 1 (not accurate): The remaining premise fails to complete the provided premise for deducing the conclusion. It may be irrelevant or inconsistent with the provided premise or conclusion, or both. \\
- 2 (somewhat accurate): The remaining premise can somewhat supplement the provided premise but is not entirely sufficient for a conclusion inference. It may require additional information for comprehensive completion, or contradicts common sense, or lacks clarity. \\
- 3 (highly accurate): The remaining premise, combined with the provided partial premise, can correctly lead to the given conclusion, and also aligns well with common sense. \\
Please first output your rating based on your general knowledge and logical reasoning, and then provide a brief explaination with no more than 100 words. \\

\text{[}Conclusion\text{]}: \{conclusion\} \\
\text{[}Partial Premise\text{]}: \{partial premise\} \\
\text{[}Remaining Premise\text{]}: \{rest premise\} \\
\text{[}Output\text{]}:
\end{tcolorbox}

\begin{tcolorbox}[colback=blue!5!white,colframe=black,width=1.0\textwidth,title={Prompt for Rating the Accuracy of Premise Generation}]
\small
You are a helpful scoring assistant. \\
Please carefully read the provided conclusion along with the specified number of facts, and rate the accuracy of candidate premise in both reaching the conclusion and containing the correct number of facts, using a scale from 1 to 3: \\
- 1 (not accurate): The premise is logically incorrect, irrelevant or contradictory for deducing the conclusion, or it contains an incorrect number of facts. \\
- 2 (somewhat accurate): The premise can partially infer the conclusion but is not entirely sufficient. It may require additional information, or contradicts common sense, or lacks clarity. \\
- 3 (highly accurate): The premise can correctly lead to the given conclusion and aligns well with common sense, and precisely contains the specified number of facts. \\
Please first output your rating based on your general knowledge and logical reasoning, and then provide a brief explaination with no more than 100 words. \\

\text{[}Fact Number\text{]}: \{fact num\} \\
\text{[}Conclusion\text{]}: \{conclusion\} \\
\text{[}Premise\text{]}: \{premise\} \\
\text{[}Output\text{]}:
\end{tcolorbox}
\caption{Prompts for rating the accuracy of three tasks.}
\label{accuracy_rating}
\end{table*}

\begin{table*}[h!]
\begin{tcolorbox}[colback=blue!5!white,colframe=black,width=1.0\textwidth,title={Prompt for Rating the Diversity of Premise Generation}]
\small
You are a helpful scoring assistant. \\
Please read the provided conclusion and multiple generated premises carefully, and rate the diversity of these premises using a scale from 1 to 3: \\
- 1 (low diversity): The premises show minimal variation, where all three premises largely repeat same perspectives with slight lexical changes. \\
- 2 (moderate diversity): The premises exhibit some degree of variation, with two out of the three premises sharing similar perspectives, expressions and fact numbers while the third presents different content. \\
- 3 (high diversity): The premises display a high level of diversity, where each premise presents distinct perspective from the others, or contains different fact numbers. \\
Please first output your rating, and then provide a brief explaination with no more than 50 words. \\

\text{[}Conclusion\text{]}: \{conclusion\} \\
\text{[}Premise\text{]}: \{premise$_1$\}, \{premise$_2$\}, \{premise$_3$\} \\
\text{[}Output\text{]}:
\end{tcolorbox}

\begin{tcolorbox}[colback=blue!5!white,colframe=black,width=1.0\textwidth,title={Prompt for Rating the Complexity of Premise Generation}]
\small
You are a helpful scoring assistant. \\
Please carefully read the provided conclusion, and rate the complexity of candidate premise considering both the number of facts it comprises and its semantic difficulty, using a scale from 1 to 3: \\
- 1 (low complexity): The premise is straightforward, incorporating no more than 3 facts with clear and easy-to-understand semantics and a simple logical structure. \\
- 2 (moderate complexity): The premise exhibits moderate complexity, which involves 4 facts and somewhat intricate semantics and a logical structure that require some thought to understand. \\
- 3 (high complexity): The premise is highly complex with more than 4 facts, which also includes complex semantics and an abstract logical structure, demanding a high level of understanding. \\
Please first output your rating based on your general knowledge and logical reasoning, and then provide a brief explaination with no more than 50 words. \\

\text{[}Conclusion\text{]}: \{conclusion\} \\
\text{[}Premise\text{]}: \{premise\} \\
\text{[}Output\text{]}:
\end{tcolorbox}

\begin{tcolorbox}[colback=blue!5!white,colframe=black,width=1.0\textwidth,title={Prompt for Rating the Abstractness of Premise Generation}]
\small
You are a helpful scoring assistant. \\
Please carefully read the provided conclusion, and rate the abstractness of objects in the candidate premise considering how broadly they can generalize to various specific instances, using a scale from 1 to 3: \\
- 1 (low abstractness): The objects in the premise are concrete and specific, making direct and clear reference to particular instances or examples, which focus on specific people, places, or tangible entities, such as Swimmer, New York, or SUV. \\
- 2 (moderate abstractness): The objects in the premise are somewhat abstract, representing a balance between specific instances and general concepts. They may pertain to fine-grained categories of people, places, or things, such as Professionals, City, or Car. \\
- 3 (high abstractness): The objects in the premise are highly abstract, focusing on coarse-grained people, places or things that are far removed from concrete instances, such as Person, Region, or Event, or general properties like Age and Height. \\
Please first output your rating based on your general knowledge and logical reasoning, and then provide a brief explaination with no more than 50 words. \\

\text{[}Conclusion\text{]}: \{conclusion\} \\
\text{[}Premise\text{]}: \{premise\} \\
\text{[}Output\text{]}:
\end{tcolorbox}

\caption{Prompts for rating the diversity, complexity and abstractness of premise generation.}
\label{diversity_complexity_rating}
\end{table*}

\subsection{Human Evaluation Templates}
\label{human_eval_template}
\begin{figure*}[!h]
    \centering
    \includegraphics[width=1.0\textwidth]{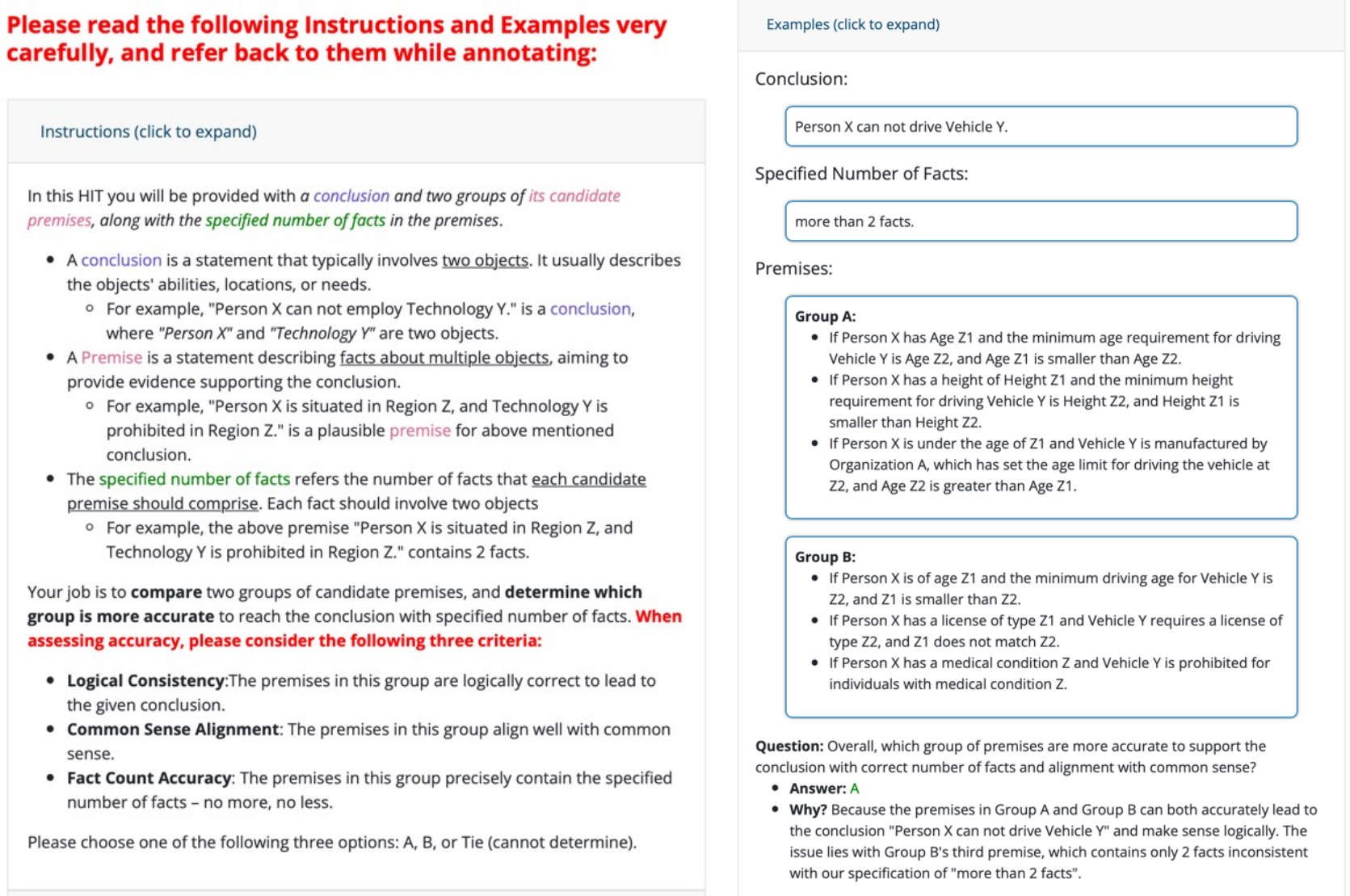}
    \caption{AMT template for human evaluation for premise generation accuracy.}
    \label{human_eval_figure}
\end{figure*}

For the human evaluation of premise generation accuracy, we qualify a new cohort of AMT annotators to pairwise compare two sets of generated premises in terms of logical consistency with the provided conclusion, alignment with common sense and the inclusion of an accurate number of facts.
The detailed template for human evaluation is shown as Figure~\ref{human_eval_figure}. 

\subsection{Downstream Reasoning Datasets}
\label{appen:downstream_reasoning_datasets}
StrategyQA and SOCIAL IQA consist of crowd-sourced questions involving reasoning of implicit logic. LINK comprises GPT-4 generated statements instantiated from abstract rules, including two subsets: head distribution statements and long-tail knowledge statements. PIQA examines operational commonsense for achieving physical goals and CSQA2.0 features adversarial commonsense examples designed to mislead AI systems. 
\end{document}